\pdfoutput=1
\documentclass[11pt]{article}

\usepackage{ACL2023}

\usepackage{times}
\usepackage{latexsym}

\usepackage[T1]{fontenc}
\usepackage[utf8]{inputenc}

\usepackage{microtype}

\usepackage{inconsolata}

\usepackage{algorithm}
\usepackage{algorithmic}
\usepackage{booktabs}
\usepackage{listings}
\definecolor{codegreen}{rgb}{0,0.6,0}
\definecolor{codegray}{rgb}{0.5,0.5,0.5}
\definecolor{codepurple}{rgb}{0.58,0,0.82}
\definecolor{backcolour}{rgb}{0.95,0.95,0.92}

\usepackage[skins]{tcolorbox}

\usepackage{newfloat}
\usepackage{listings}
\usepackage{enumitem}

\title{How Good Are LLMs at Processing Tool Outputs?}

\author{
  Kiran Kate$^{1*}$,
  Yara Rizk$^{1*}$,
  Poulami Ghosh$^2$, 
  Ashu Gulati$^3$, \\
  {\bf
  Tathagata Chakraborti$^1$,
  Zidane Wright$^1$,
  Mayank Agarwal$^1$} \\
  $^1$IBM Research, USA  \hspace{10mm} $^2$Persistent Systems \hspace{10mm} $^3$Grab \\
  \normalsize{kakate@us.ibm.com, yara.rizk@ibm.com} \\
}

\begin{document}
\maketitle
\begin{abstract}
Most realistic task automation problems require large language models (LLMs) to call tools, which often return complex JSON responses. These responses must be further processed to derive the information necessary for task completion. The ability of LLMs to do so is under-studied. In this paper, we study the tool response processing task and LLMs' abilities to process structured (JSON) responses. We created a dataset for this task, and evaluated 15 open and closed weight models using multiple prompting approaches\footnote{\url{https://github.com/LongFuncEval/toolJSONprocessing}}. 
Our results show that JSON processing remains a difficult task even for frontier models across multiple prompting strategies. The optimal response processing strategy depends on both the nature and size of the tool outputs, as well as the complexity of the required reasoning. Variations in processing approaches can lead to performance differences ranging from 3\% to 50\%.
%Our analysis highlights the importance of understanding the characteristics of tool-generated responses when designing multi-turn tool calling systems, and underscores the need for selecting an appropriate response processing strategy.
%We conducted detailed analysis of these approaches and conclude that it is critical to understand the characteristics of tool responses when designing a multi-turn tool calling system and choose a tool response processing approach 
% ,
% especially when the task becomes more complex.
\end{abstract}

\section{Introduction}
\label{sec:intro}
\begin{figure}[tb]
    \centering
    \includegraphics[width=\linewidth]{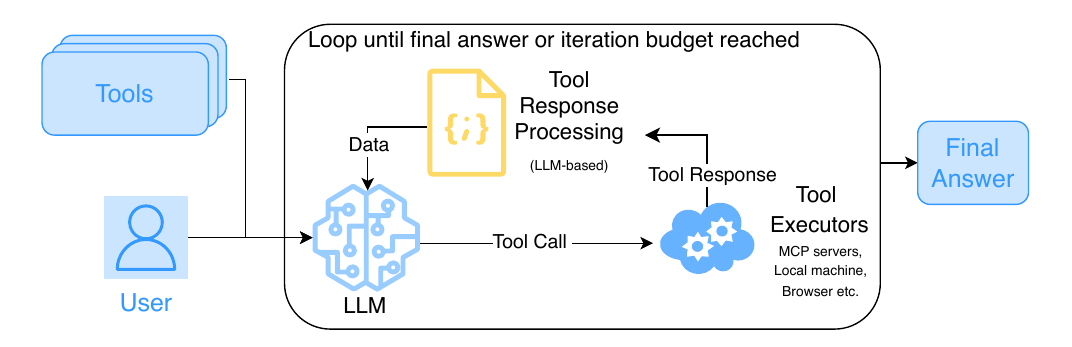}
    \caption{A typical multi-turn tool use pipeline. Tool response processing is shown here as an explicit step to highlight its importance. This paper studies LLM-based approaches to tool response processing and measures the correctness of output "Data" from the processing step by creating a handcrafted evaluation dataset for it.}
    \label{fig:flow}
\end{figure}

Tool-augmented large language models (LLMs)~\citep{qin2023toolllm, li2023api, xu2023gentopia, lu2025toolfive, shen2024llm} can perform tasks beyond their parametric knowledge by calling external tools. 
A typical multi-turn tool-calling scenario is shown in Figure \ref{fig:flow}. Given a user's request, an LLM may need to use multiple tools in sequence to address it. Each tool call generated by the LLM is executed and the tool returns data that contains information useful for either responding to the user or for performing the next steps leading to the final answer. Real-world enterprise tools may be existing REST APIs which return large JSON objects as their response~\cite{qintoolllm, zhong2025complexfuncbench}. The process of extracting useful information from their response, then, becomes non-trivial.

A common approach is to simply append the tool response to the LLM context~\citep{yao2023reactsynergizingreasoningacting}. This works well for scenarios with small and simple tool outputs~\citep{li2023api, liu2024apigenautomatedpipelinegenerating, basu2025nestful}. For real-world APIs with long and complex outputs, the response is either added as is or is pruned to fit in the context~\citep{qin2023toolllm, zhong2025complexfuncbench}. Sometimes, LLMs are used to summarize the tool output~\citep{aksitov2023rest} or to generate code to process it~\citep{song2023restgpt}.

When building real-world agentic systems, it is crucial to understand which of these tool response processing approaches works best. This analysis remains a gap in the literature despite the existence of many works focused on understanding LLMs' tool selection and end-to-end task execution abilities. 
In this paper, we address this gap by isolating the tool response processing task, creating an evaluation dataset, and analyzing LLMs' performance.

The task is formulated as question-answering (QA) on structured tool responses. Specifically, we consider the JSON response format which is popular among real-world tools~\citep{qintoolllm}. 
We manually create the QA dataset based on real tool responses, obtained from RapidAPI\footnote{\url{https://rapidapi.com/}}. 
We evaluate 15 high performing LLMs across model families, model sizes, reasoning abilities and context lengths. 
We compare different prompting techniques including prompting the model to generate the final answer versus to generate Python code to parse the JSON, utilizing chain of thought (CoT) reasoning, and adding various arguments (response, schema, etc.) to the prompt. 
We structure our analysis across several key dimensions: response length, prompting approach, and question type. 

Our experimentation and analysis led to three main conclusions: 1) JSON processing is a non-trivial task even for frontier models (GPT-4o highest accuracy is 77\%), 2) models can handle this task better by generating JSON parsing code instead of deriving the answer by looking at the JSON when the processing involves reasoning over the data (ranging from +3\% to +50\%), 3) including a schema of the structured data in the prompt improves performance (up to +12\%), and 4) including even a condensed version of the response is better than excluding the response entirely when it doesn't fit in the model's context window. Section~\ref{sec:findings} provides detailed results and ablation studies. 

Our contributions are as follows: 
\begin{itemize}[noitemsep, topsep=0pt]
    \item We formulated tool response processing as a QA task and created an evaluation dataset;
    \item We evaluated 15 open and closed weight models on this task with different processing approaches focusing on 4 main control points;
    \item We performed an in-depth analysis of LLM performance and derived practical insights to inform real-world applications.
\end{itemize}

\section{Tool Response Processing}
\label{sec:tool_response_processing}
The ability of LLMs to extract and/or derive useful information from tool responses is crucial to the success of end-to-end agentic systems. As shown in Figure \ref{fig:flow}, our goal is to isolate the tool response processing component within an agent and evaluate the state of the art approaches for that task.

\subsection{Dataset Creation}
\label{sec:data}
The task is formulated as a QA task on tool responses. The dataset consists of triplets (tool response, natural language question, answer) where the tool response is in JSON format and the answer to the question is based on the data in the response. The queries are categorized into one of three types: extractive (returns a value in the JSON given a key), filtering (returns multiple entries corresponding to filtering criteria) and aggregation (combines multiple entries by performing an aggregation operation). Figure~\ref{fig:processing} shows an example of these three types of questions (with additional examples in Figures \ref{fig:extractive_templates}, \ref{fig:filtering_templates}, \ref{fig:aggregation_templates} in the Appendix). 

To perform our study, we collected multiple responses from six REST API endpoints from RapidAPI, covering domains such as hotel booking, car rentals, flight seats, SEC filings, and product search. For the travel related endpoints, we extract the API responses from the ComplexFuncBench dataset~\citep{zhong2025complexfuncbench}. We manually reviewed those responses to understand the JSON structure, created question templates and wrote Python code to retrieve the ground truth answers. 

Our dataset contains 1298 QA-samples (Table \ref{tab:data_stats}), with 580 extractive (derived from 14 templates), 394 filtering (18 templates), and 324 aggregation (18 templates) questions. The JSON responses varied in length (averaging 24,000 to 74,000 characters) and complexity. 

\begin{table*}[ht] % Numbers in the table were updated Sep. 22, 2025
    \centering
    \small
    \begin{tabular}{l c c c c c}
    \toprule
         Endpoint & Total Samples & Extr. & Filt. & Aggr. & Avg. Response Length (\#Characters) \\
         \midrule
         Booking - Get Hotels & 13 & 6 & 5 & 2 & 25,008\\
         Booking - Search Car Rentals & 87 & 34 & 34 & 19 & 24,048\\
         Booking - Get Seat Map & 184 & 29 & 89 & 66 & 53,698\\
         Real Products - Shoes & 228 & 75 & 75 & 78 & 55,968\\
         SEC-Filings & 267 & 84 & 98 & 85 & 74,319\\
         Booking - Get Rooms & 519 & 352 & 93 & 74 & 65,341\\
         \midrule
         Total & 1298 & 580 & 394 & 324 & - \\
         \bottomrule
    \end{tabular}
    \caption{Dataset statistics including sub-type of queries (extractive, filtering or aggregation) per endpoint}
    \label{tab:data_stats}
\end{table*}

\subsection{Common Approaches}

\begin{figure}[tb]
    \centering
    \includegraphics[width=1\linewidth]{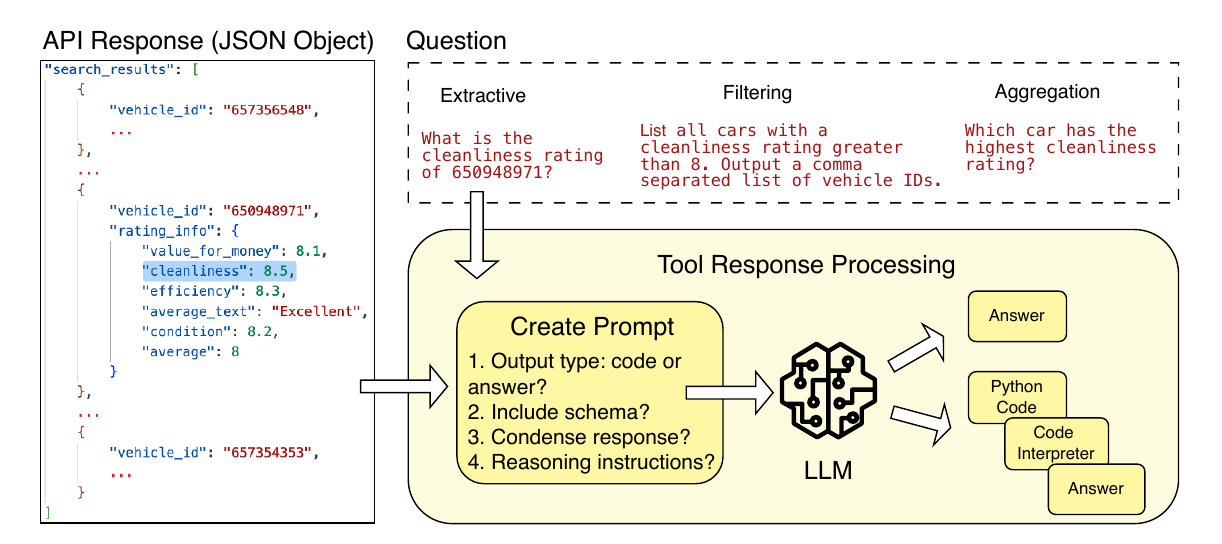}
    \caption{Tool response processing approaches based prompting an LLM and varying four control points: the type of model output (generate code or the final answer); whether the schema of the API response is included in the prompt; whether the full response is included, condensed or excluded; and whether any reasoning instructions are included (e.g, chain-of-thought type prompts).}
    \label{fig:processing}
\end{figure}

We consider multiple processing approaches from the literature \cite{aksitov2023rest, song2023restgpt},  all relying on zero-shot prompting of the LLM. The variations in the approaches, captured in Figure \ref{fig:processing}, stem from four control points. 
(1) \textit{Model output type}: we can instruct the model to either generate the answer from the response (referred to as \textit{answer generation})  or generate Python code to parse the JSON response to obtain the answer (referred to as \textit{code generation}). The latter approach is based on the observation that LLMs are very familiar with the JSON format and are good at generating code. Such generated code is then executed to get the answer.
(2) \textit{Prompt type}: we consider two main templates, one with reasoning instructions (CoT \cite{wei2022chain}), and one without. 
(3) \textit{Schema}: we evaluate the prompt template with and without the JSON schema of the response. 
(4) \textit{Tool response}: for code generation with schema setups, we consider the impact of including the response (whether in full or a reduced version) vs. excluding it. This enables us to study cases where the response doesn't fit in the context of an LLM. 

We include the example prompt templates in the Appendix \ref{sec:prompts}. 
We do not consider approaches based on text summarization or RAG (retrieval augmented generation). During our preliminary testing of such approaches, we observed information loss (particularly in the relationships derived from the structure), reducing our confidence in the suitability of these approaches for JSON processing. 

\section{Experimental Setup}
\label{sec:setup}
\subsection{Metrics} \label{sec:metrics}
%The literature often relies on one of four main metrics.
In the various research questions (RQs) we define, we evaluate performance on the following three metrics: 1) \textit{Exact match} compares the predicted answer to the ground truth answer with some relaxations such as white spaces, casing, order of elements in lists, duplicates, rounding errors. This metric is often considered the strictest as it penalizes the model for verbose outputs and semantic variations.
% for format and semantic variations. However, to account for some reasonable variations, we compare ground truths to the predicted answer based on the data type of the ground truth. When the data type is string, we ignore white spaces and casing. For answers that are lists, we rely on an unordered list match which ignores order, duplicates, and minor formatting discrepancies. For numeric answers, we apply an approximate number match to account for differences in rounding between the model's answer and the ground truth.
2) \textit{Contains} is a disjunction between \textit{exact match} and recall based accuracy (which checks whether the ground truth exists within the predicted answer); this gives us an upper bound on programmatic match. 
3) \textit{LLM as a judge} employs an LLM to compare the predictions to the ground truth, mostly focusing on their semantic equivalence; this metric is lenient and has a high false positive rate. We used Llama-3-3-70b-instruct as the judge (see Appendix~\ref{sec:prompts} for the prompt).

\subsection{Models}
We evaluate 13 open weight high performing models (see Table \ref{tab:models} in the Appendix) with varying sizes (8B to 480B+ active parameters) and 2 closed weight models: GPT-4o and Claude-4-Sonnet. Inference is done using greedy decoding with the temperature set to 0 and generate a maximum of a 1000 tokens; all our prompts are zero-shot (see Appendix \ref{sec:prompts}). Since the tool responses in our dataset are large, we only consider models with 65K context or greater.

\section{Study Findings}
\label{sec:findings}
% We analyze the results of our study by formulating multiple research questions (RQ). 

\subsection{Impact of Response Length (RQ1)}
First, we evaluated whether long responses make the task harder. We designed a controlled setup where we can create a JSON object of a given size from the dataset mentioned in Section \ref{sec:data} 
by merging multiple JSON records from the API responses. The size is indicated by the number of tokens (as per Llama-3-1-70b-instruct tokenizer) and exponentially increases from 10K to 80K tokens.

\begin{figure}[tb]
    \centering
    \includegraphics[width=1\linewidth]{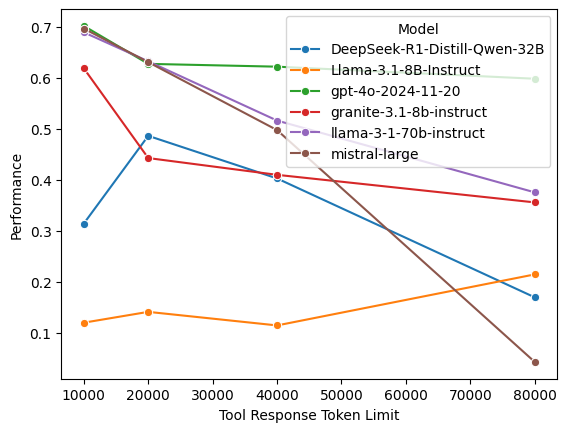}
    \caption{The exact match accuracy across models with increasing JSON size shows that models struggle with longer responses.}
    \label{fig:rq1_plot}
\end{figure}

Figure~\ref{fig:rq1_plot} shows the performance of LLMs based on the \textit{exact match} metric as the JSON response token size is varied. Performance generally declines as response size increases, with GPT-4o experiencing a 7\% drop and Mistral-large showing the steepest decline at 91\%. LLMs are known to have limitations in handling longer contexts effectively, even for simpler tasks~\citep{niah, hsieh2024ruler}. This finding confirms the challenge of dealing with tool responses as they become larger.

\begin{center}
\begin{tcolorbox}[skin=widget,
boxrule=1mm,
coltitle=black,
colframe=yellow!45!white,
colback=yellow!15!white,
width=(\linewidth),center,before=\hfill,after=\hfill]
\textbf{RQ1}: Does the length of the tool response impact the performance of LLMs? \\
\textbf{Insight}: LLM performance degrades as the length of the tool responses increases.  
\end{tcolorbox}
\end{center}

\subsection{Impact of Answer Location (RQ2)}
Long context evaluation of LLMs also includes the impact of the position of the key in retrieval tasks. Behaviors such as lost in the middle~\cite{liu-etal-2024-lost} 
%where the models don't perform well when the answer is in the middle of the prompt 
and recency bias~\cite{peysakhovich2023attentionsortingcombatsrecency} 
% where the models perform well when the answer is close to the end of the prompt 
have been reported on other tasks. 

To evaluate the impact of the answer's position on LLM performance, we note that API responses are typically a list of records of some entity type (e.g., flights in flight search). The position of the answer record in the tool response from the triplets (tool response, natural language query, answer) in our dataset can be varied. For the 80K token size setting of RQ1, we vary the position of the answer record (p) from 1 to 8. 

\begin{figure}[tb]
    \centering
    \includegraphics[width=\linewidth]{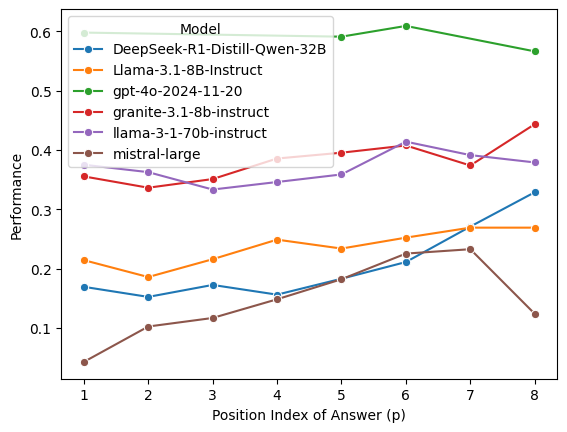}
    \caption{The exact match accuracy of models is impacted by the response's position indicating that models suffer from recency bias.}
    \label{fig:resp_position}
\end{figure}

Results, in Figure \ref{fig:resp_position}, indicate that recency bias was observed in answer position in the JSON (higher values of p denote more recent information). Some models suffered more than others, the performance variation based on position
ranges from 5\% (GPT-4o) to 75\% (Mistral-large).

\begin{center}
\begin{tcolorbox}[skin=widget,
boxrule=1mm,
coltitle=black,
colframe=yellow!45!white,
colback=yellow!15!white,
width=(\linewidth),center,before=\hfill,after=\hfill]
\textbf{RQ2}: Does the location of the answer impact the performance of LLMs? \\
\textbf{Insight}: LLMs exhibited recency bias.
\end{tcolorbox}
\end{center}

\subsection{Impact of Model Output Type (RQ3)}
In the remaining subsections, we evaluate the performance of various prompting approaches and models on the task of processing JSON tool responses. First, we study the impact of the model output type, two possibilities specifically. The first option is to prompt the model to directly process the JSON response and output the answer (termed ``answer'' in the Figures). The second option is to prompt the model to generate Python code that is then executed to parse the JSON response and return the answer (termed ``code'' in the figures). 

\begin{figure*}
    \centering
    \includegraphics[width=1\linewidth]{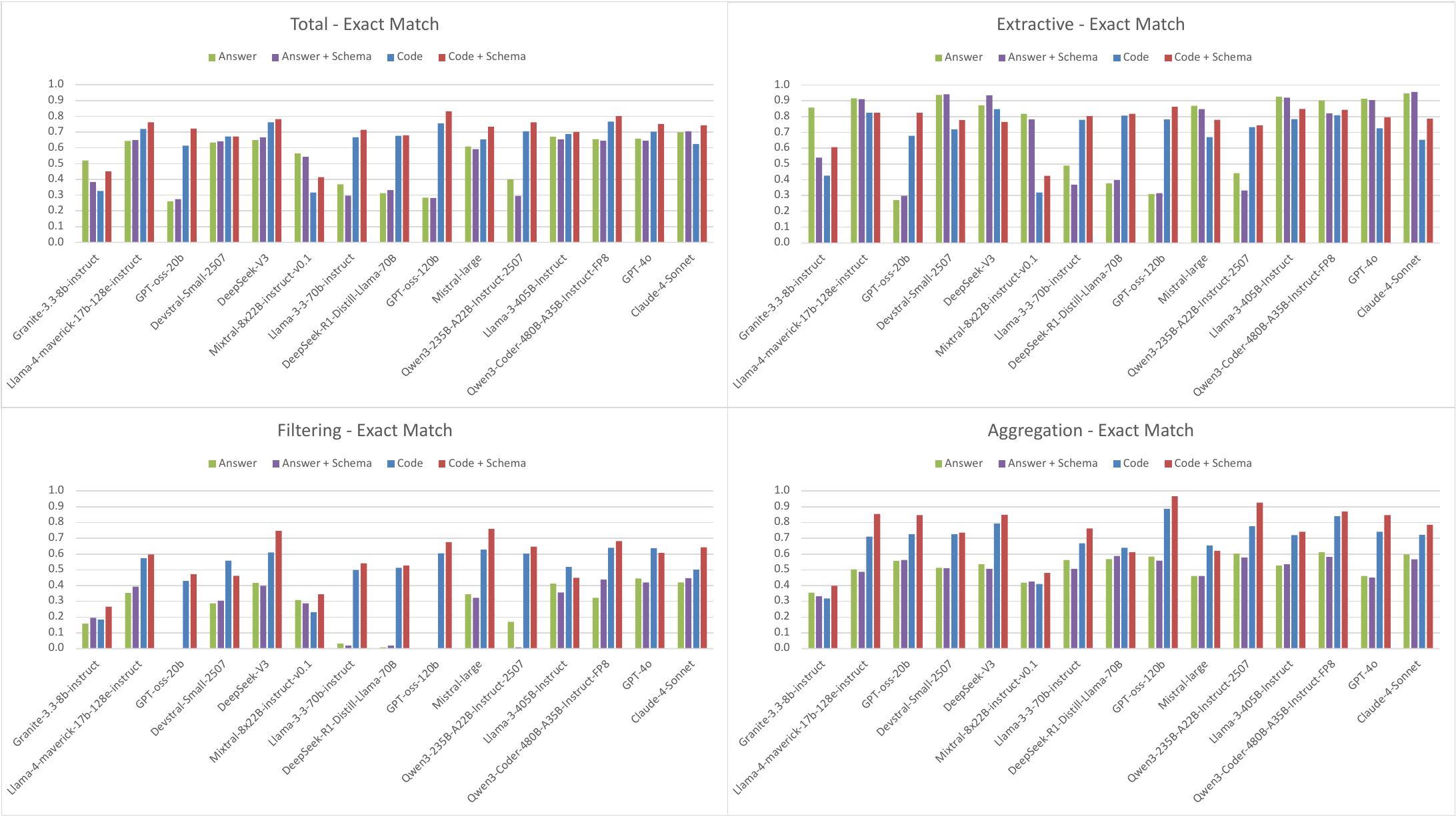}
    \caption{The exact match accuracy comparing model output types shows that generating code is more accurate for filtering and aggregation questions. Adding the response schema generally improves performance across models and question categories. Models are ordered from left to right in increasing number of active parameters.}
    \label{fig:rq3_4_exact_match}
\end{figure*}

% code vs. direct
The ``answer'' and ``code'' settings in Figure \ref{fig:rq3_4_exact_match} capture the results of this experiment. 
% assume that the full response is included in the prompt (as opposed to a condensed version) without a schema and no reasoning specific prompting (i.e., without chain of thought). 
% Exact match accuracy on 15 models is reported on the three QA categories of extractive, filtering, and aggregation. 
As seen from the results, the best performance out of these two options is close to 76\%, underscoring the complexity of the task. The performance difference between the two output types ranges from 2\% to 47\% confirming that output type is an important factor to take into consideration and models such as GPT-oss-120b are able to work with one much better than the other.
In general, the total accuracy when generating code is better for 13 out of 15 models. However, for extractive questions, answer generation is better for 10 out of 15 models including a strong code model such as Claude-4-Sonnet. This shows that the choice of output type is not straight forward and it depends on the type of processing required. As the processing needs become more complex (filtering and aggregation), code generation is more accurate for all the models.

\begin{figure}[tb]
    \centering
    \includegraphics[width=1\linewidth]{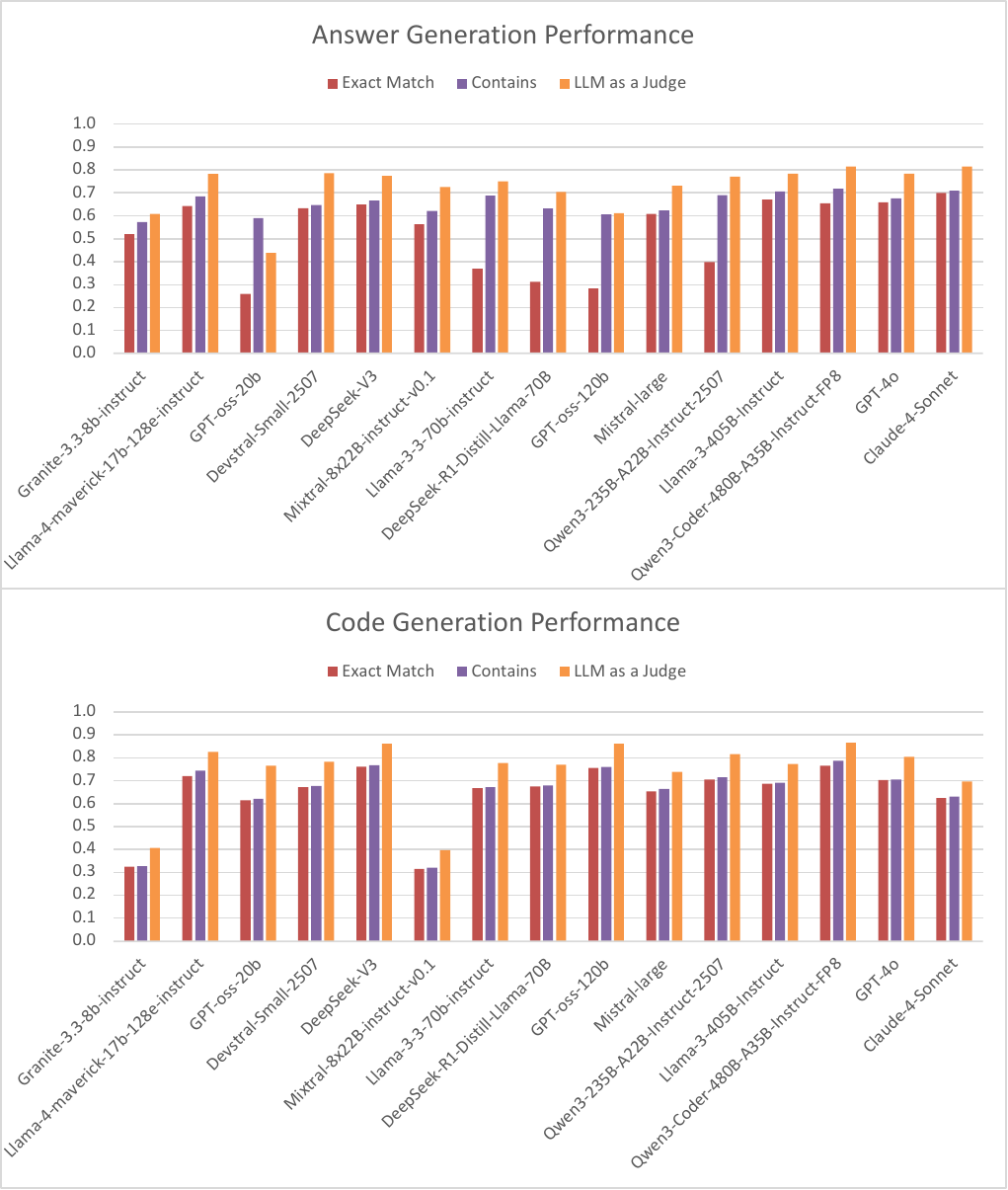}
    \caption{Performance comparison across different metrics for all 15 models when generating the answer (top) and generating the code (bottom) show that exact match is the strictest and LLM as a judge is the most lenient.}
    \label{fig:rq3_metrics}
\end{figure}

Furthermore, multi-turn conversations increase in length, answer generation based approaches will cause the context to grow much more quickly than code generation based approaches. As such, code generation may be more suitable for longer conversations. 
% metrics: exact match vs. contains vs. LLM as a judge
Figure \ref{fig:rq3_metrics} shows the performance of both settings when considering the three metrics: exact match, contains, and LLM as a judge (as defined in section \ref{sec:metrics}). Exact match is a strict metric that penalizes models for any formatting discrepancies. The contains metric serves as an upper bound for the performance of specific approaches. 

We observe that when generating the answer, unlike when generating code, the model is more likely to output superfluous text beyond the answer (as indicated in the gap between exact match and contains). This makes it challenging to leverage answer generation in scenarios with strict formatting requirements, such as when the output of one tool must be passed as the input to the next tool. LLM as a judge is often the most lenient of all these metrics. As can be seen, even with the most lenient semantic comparison, performance remains much below 90\% across the board.
% , but some anomalies existed (e.g., GPT-OSS-20B). 
 
\begin{center}
\begin{tcolorbox}[skin=widget,
boxrule=1mm,
coltitle=black,
colframe=yellow!45!white,
colback=yellow!15!white,
width=(\linewidth),center,before=\hfill,after=\hfill]
\textbf{RQ3}: How does the model output type impact the LLMs' performance? \\
\textbf{Insight}: Code generation performs better for filtering and aggregation while answer generation is better for extractive.
\end{tcolorbox}
\end{center}

\subsection{Impact of Response Schema (RQ4)}
% Setup description
We build on the setup from RQ3 and add the API response JSON schema to determine how the schema impacts the performance of both answer and code generation approaches. The schema captures the structure of the JSON response and information such as data types, default values, and enum values of individual data fields. This experiment is to understand if models are able to use this extra knowledge and perform better. We generate the schema from an API response (details in Appendix). 

% w/ vs w/o Schema + DP vs CG w/ Schema
We can see from Figure \ref{fig:rq3_4_exact_match} that both prompting for the answer and code generally benefit from adding the schema information (all 15 for code generation, 7 out of 15 for answer generation). The performance difference is more significant for code generation than answer generation. 
%Furthermore, code generation with schema always outperforms answer generation with schema on the filtering and aggregation samples. 

% DPS -> CGS results
Table \ref{tab:DPS2CGS} 
shows how the distribution of correct vs. incorrect output varies when we change the setup from answer generation with schema to code generation with schema for GPT-4o (similar distribution patterns were observed for other models). Ideally, we want all the samples that were false to become true ("F to T" column) and none of the true become false ("T to F" column). Results show that the filtering and aggregation samples, which were problematic for answer generation, were mostly positively impacted by the change. The extractive category has a significant number of samples going from true to false which is not ideal. Thus, if it were possible to identify a query as extractive, filtering or aggregation, we can route it to the appropriate approach and attain the best of both worlds. 

% \begin{table*}[tb]
% \centering
% \begin{tabular}{l l c c c c}
% \toprule
% Model     & Task type   & True $\to$ True & True $\to$ False    & False $\to$ True  & False $\to$ False  \\ \midrule
% gpt-4o          & Extractive (580)  &    446                 &         \color{red}{79}                  & \color{green}{6} & 49   \\
% gpt-4o          & Filtering (394)  &    123                  &        \color{red}{31}                   &     \color{green}{102}                   &  127 \\

% gpt-4o          & Aggregation (324) &       138               &       \color{red}{8}                    &         \color{green}{128}               &   50 \\
% gpt-4o          & Total (1298)       &      718               &      \color{red}{118}                    &      \color{green}{236}                  & 226  \\
% \bottomrule
% \end{tabular}
% \caption{Change in distribution for samples when switching from answer generation to code generation. Red is used to denote that lower is better and green that higher is better.}\label{tab:DPS2CGS}
% \end{table*}

% \begin{figure}[tb]
%     \centering
%     \includegraphics[width=\linewidth]{figures/rq4_dps2cgs.png}
%     \caption{Change in distribution for samples when switching from answer generation to code generation; lower is better for True $\to$ False and higher is better for False $\to$ True}
%     \label{fig:DPS2CGS}
% \end{figure}

\begin{table}[tb]
    \centering
    \begin{tabular}{l p{1cm} p{1.5cm} p{1.5cm}}
    \toprule
        Task type & No Change & T to F ($\downarrow$) & F to T ($\uparrow$) \\ \midrule
        Extractive & 0.85 & 0.14 & 0.01  \\ 
        Filtering & 0.66 & 0.08 & 0.26  \\ 
        Aggregation & 0.58 & 0.02 & 0.40  \\ 
        \hline
        Total & 0.73 & 0.09 & 0.18 \\ 
    \bottomrule
    \end{tabular}
    \caption{Change in distribution for samples when switching from answer to code generation (with schema using GPT-4o); lower ($\downarrow$) is better for True (T) to False (F) and higher ($\uparrow$) is better for False to True}
    \label{tab:DPS2CGS}
\end{table}

% \begin{figure}
%     \centering
%     \includegraphics[width=\linewidth]{figures/rq4_sankey.png}
%     \caption{Sankey diagram showing the change in distribution for all the samples when switching from answer generation to code generation for GPT-4o; lower is better for True $\to$ False (red) and higher is better for False $\to$ True (green)}
%     \label{fig:DPS2CGS}
% \end{figure}

\begin{center}
\begin{tcolorbox}[skin=widget,
boxrule=1mm,
coltitle=black,
colframe=yellow!45!white,
colback=yellow!15!white,
width=(\linewidth),center,before=\hfill,after=\hfill]
\textbf{RQ4}: Does adding response schema impact LLM performance? \\
\textbf{Insight}: LLM performance improves with addition of schema. Higher improvement for code than answer generation.
\end{tcolorbox}
\end{center}   

\subsubsection{Classifying Queries}
We briefly digress to evaluate whether we can accurately predict whether a question of type extractive, filtering or aggregation. 
% Since we have that information in our dataset, then it is easy to leverage our dataset without additional changes to test this hypothesis. 
We zero-shot prompted an LLM giving it the question and the API response and ask it to predict the category (see Appendix \ref{sec:prompts} for the prompt). 

Results in Table \ref{tab:classifier_res} show that some LLMs can reasonably accurately predict the class of a query. 
% However, there is room for improvement, especially if we refine the wording of the prompt or include in-context examples. 
One caveat is that this dataset is small, when considering the question templates to category pairs (50 total unique templates from which 1200+ samples are derived). This may not be representative enough of the space of processing needs and categories that would be encountered in realistic settings. 
% However, it gives us an idea of the feasibility of such approach.

Maintaining dual processing pipelines significantly increases system complexity. Practitioners must carefully balance this engineering overhead against potential performance gains. As models continue to advance, they will increasingly handle such queries directly, reducing the need for intricate supporting systems. However, certain applications may still benefit from this approach, particularly when constrained by model choices (e.g., open-source models, which remain less capable than proprietary frontier models, often require system-level enhancements to achieve acceptable performance).

\begin{table}[tb]
    \centering
    \small
    \begin{tabular}{l c c c c}
    \toprule
         Model &  Total & Extr. & Filt. & Aggr. \\
         \midrule
         DeepSeekV3 & 0.80 & 0.69 & 0.95 & 0.84 \\
         GPT-4o & 0.77 & 0.69 & 0.70 & 0.99 \\
         GPT-oss-120B & 0.76 & 0.62 & 0.78 & 0.99 \\
         Llama-3-3-70b-instr. & 0.62 & 0.69 & 0.73 & 0.38 \\
         \bottomrule
    \end{tabular}
    \caption{Models' accuracy in predicting a question's category (extractive vs. filtering vs. aggregation). DeepSeekV3 is the most accurate overall.}
    \label{tab:classifier_res}
\end{table}

\subsection{Impact of the JSON Response (RQ5)}
Next, we focus on the code generation setup and study the impact of including the API response in the prompt. Specifically, we compare three setups: using the full response (which is not always possible when the response length exceeds the context length), reducing the API response before adding it to the prompt (using a heuristic algorithm described in the Appendix), and not using the response at all (as one could argue the response schema has most of the information needed to generate code).  

\begin{figure}[tb]
    \centering
    \includegraphics[width=1\linewidth]{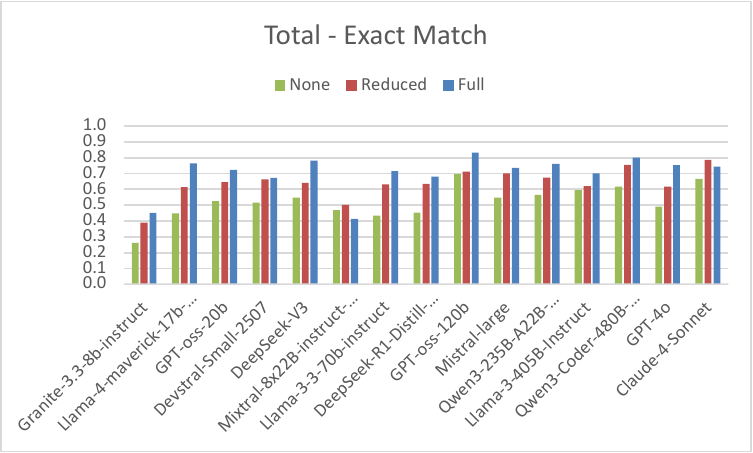}
    \caption{The performance (exact match reported) of the code generation approach on all models is highest when including the full response. However, when that is not possible due to context window constraints, including a reduced version of the response is better than excluding it entirely. It is not possible to perform this experiment on answer generation since the full response must be in the prompt for that setting to work. }
    \label{fig:rq5_total_exact_match}
\end{figure}

% CG only w/ schema: w/o vs. condensed vs. full
As seen in Figure \ref{fig:rq5_total_exact_match}, passing the full response achieved the highest accuracy across all models. However, when that is not possible due to context length constraints, using a condensed response is better than not adding a response at all. Upon investigation, we observed that the schema does not always include all the information about the values of the keys that may require the model to include specific Python code to properly parse the API response (e.g., ``price'' is a string and all the values in the response start with ``USD'').  

\begin{figure}[tb]
    \centering
    \includegraphics[width=1\linewidth]{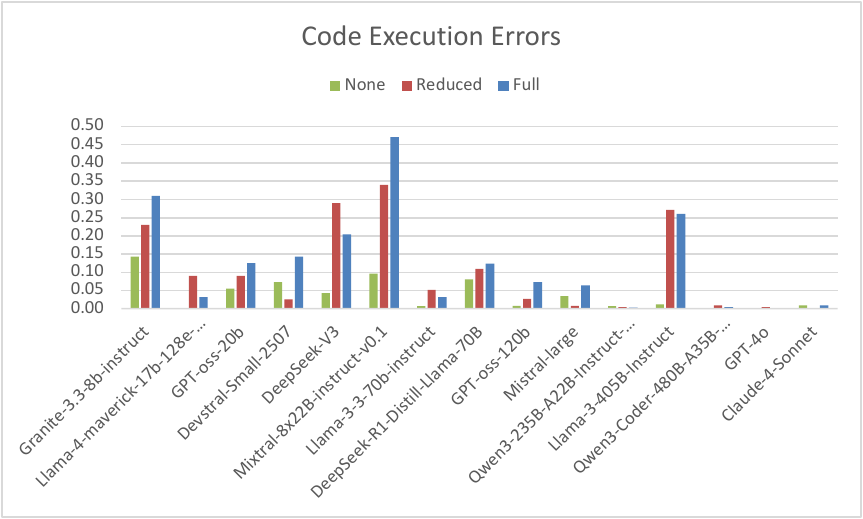}
    \caption{Code execution error rates for the code generation approach indicate that some models are more likely to fail at generating executable code than others. GPT-4o and Claude-4-Sonnet have almost 0\% errors indicating that their incorrect answers on some samples can be attributed to semantic understanding of the task or related issues.}
    \label{fig:rq5_code_exec}
\end{figure}

% code failures
Figure \ref{fig:rq5_code_exec} shows the percentage of wrong answers attributed to code failures. This allows us to understand whether the code based approaches are failing at the code execution level or at the logical level. Most models are able to generate executable code, especially the more capable models like GPT-4o and Claude-4-Sonnet. However, as the context grows, so too does the probability of the wrong answers being caused by failing code.  

\begin{center}
\begin{tcolorbox}[skin=widget,
boxrule=1mm,
coltitle=black,
colframe=yellow!45!white,
colback=yellow!15!white,
width=(\linewidth),center,before=\hfill,after=\hfill]
\textbf{RQ5}: How does the JSON response impact code generation? \\
\textbf{Insight}: Adding the full JSON response, whenever the context size permits it, will lead to the best possible performance across most model families. 
\end{tcolorbox}
\end{center}

\subsection{Impact of Reasoning Prompt (RQ6)}
Another control point is the prompt we pass to the models, particularly related to whether the prompt includes reasoning-based instructions. Since some questions, especially in the filtering and aggregation categories, may require a level of reasoning for the LLM to determine the answer, we investigated the impact of using a CoT prompt.

% - DPS vs. CGS for CoT
% - w/ vs. w/o CoT for both DPS and CGS
% - per task type (extr., filt., agg.)

\begin{figure}[tb]
    \centering
    \includegraphics[width=1\linewidth]
    {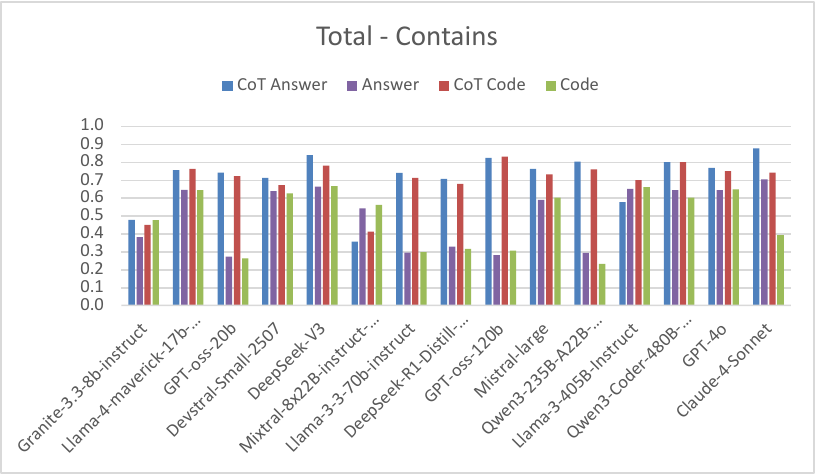}
    \caption{Performance results (contains metric) of the 15 models on reasoning-based prompting (CoT vs. none) show that both answer and code generation benefit form CoT when questions are filtering or aggregation. }
    \label{fig:rq6_contains_total}
\end{figure}

Figure \ref{fig:rq6_contains_total} reports the results for the 15 models we tested; all models are either confirmed or speculated to have been trained on reasoning traces. We also focus on the contains metric since reasoning instructions will cause models to output superfluous text that will disadvantage the exact match metric. Results show that the benefits of CoT depend on the model used. Breaking down the results further based on the type of question (reported in the Appendix) provide a better understanding of CoT's impact: extractive doesn't benefit but filtering and aggregation do. We also observe that combining reasoning with answer generation can be a good substitute when code generation is not possible. 

\begin{center}
\begin{tcolorbox}[skin=widget,
boxrule=1mm,
coltitle=black,
colframe=yellow!45!white,
colback=yellow!15!white,
width=(\linewidth),center,before=\hfill,after=\hfill]
\textbf{RQ6}: Does CoT impact performance? \\
\textbf{Insight}: CoT improves LLM performance on filtering and aggregation questions which require more reasoning.
\end{tcolorbox}
\end{center}

\subsection{Error Analysis (RQ7)}
\label{subsec:rq8}
\subsubsection{Qualitative Analysis}
To understand the kinds of errors made in both answer and code generation setups, we manually annotated 50 failures from GPT-4o and Llama-3-3-70b-instruct. Most of the errors were in 4 categories: 1) misinterpretation of the JSON structure, 2) semantic ambiguity, 3) format misalignment, and 4) code execution failure.

Misinterpretation of the JSON structure led to the model picking a value for the right key but from a wrong record. For example, when asked for area of a hotel room given the type of room, the model answered with area of a different room type. 

Semantic ambiguity is common for cases when the JSON has multiple keys that have similar meaning. The hotel room availability endpoint response has three semantically similar keys: ``name'', ``room\_name'', and ``name\_without\_policy''. Such noise is representative of real-world APIs and existing tool evaluation datasets manually remove such ambiguity~\cite{zhong2025complexfuncbench}.

Format misalignment is when models fail to follow a formatting instruction from the prompt. For example, when asked to return just the value for a key, the model returns a small JSON output with the key and value. Many of the questions in the dataset contain explicit formatting instructions.
Frontier models such as GPT-4o make less formatting errors compared to Llama-3-3-70b-instruct.

Common reasons for code execution failure were wrong data type assumptions and incorrect JSON traversal. We expect information such as JSON schema to fix some of these errors and it can be confirmed from Figure~\ref{fig:rq3_4_exact_match}.

\subsubsection{Simplifying the JSON}
To validate our observation that complex and large JSON structures were causing a lot of LLM errors, we simplify the JSON using what we call an oracle projection. Since we have the ground truth code to generate the ground truth answer, we use that code to generate a simplified JSON structure. That projection (for both the schema and the response) is then passed to the LLMs when doing answer generation and code generation.

\begin{figure}[tb]
    \centering
    \includegraphics[width=\linewidth]{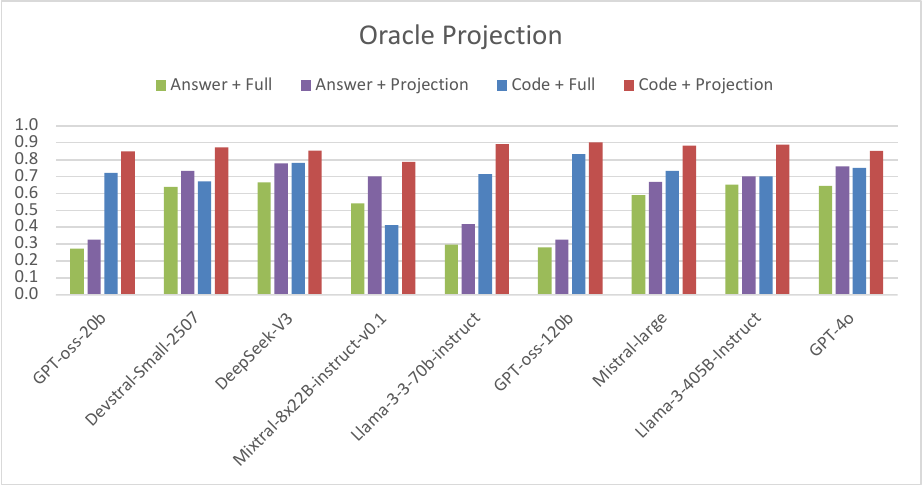}
    \caption{Simplifying the JSON using an approach that ensures the presence of the ground truth answer validates the hypothesis that complex JSON structures are causing LLMs to make mistakes in processing the responses.}
    \label{fig:rq7_cf}
\end{figure}

In Figure \ref{fig:rq7_cf}, we compare the performance of a subset of the LLM models on the projected schema and response vs. the full schema and response. We see that the simplified JSON does in fact lead to better performance across all models for both answer and code generation settings. 

The projection algorithm we used assumes the ground truth is known which is not realistic. As such this projection is an oracle and provides an upper bound on the performance gains we can achieve if we were able to simplify the JSON structure before prompting the LLM. Furthermore, the average number of characters when using full responses was 12x (60,968) higher than the average number of characters when using projected responses (5,056). Therefore, if we can simplify the response, we would also be able to reduce the inference cost while achieving higher accuracy (between 8 and 38 point improvement depending on the model).  

% \subsubsection{What if we wrote custom parsers - Format Alignment}

\begin{center}
\begin{tcolorbox}[skin=widget,
boxrule=1mm,
coltitle=black,
colframe=yellow!45!white,
colback=yellow!15!white,
width=(\linewidth),center,before=\hfill,after=\hfill]
\textbf{RQ7}: What are the common types errors observed when LLMs fail? \\
\textbf{Insight}: The complexity of the JSON structure, output format and semantic ambiguity are the top three observed errors.
\end{tcolorbox}
\end{center}

\section{Related Work}

\subsection{Tool Calling Evaluation Datasets}
The literature has many datasets created to study tool calling functionality in LLMs and agentic systems. Most focused on evaluating LLMs' ability to formulate tool calls \cite{basu2025nestful, patilberkeley, li2023api}. 
Others focused on the end-to-end task execution of agents with some data samples included tool calling \cite{trivedi2024appworld, zhou2023webarena, zhong2025complexfuncbench}. None of these datasets enable the detailed evaluation of the tool response processing step. 

% Since we model the tool response processing task as a QA problem, we considered existing QA datasets  \cite{wang2022modern}. They mostly process unstructured text (documents, web pages, etc.) to determine the answer. Some perform QA over tables \cite{chenopen} and graphs \cite{banerjee2023dblp}; however, JSON objects do have peculiarities that vary from tables or graph. Hence, we could not find any existing datasets to perform an evaluation of JSON response processing.

\subsection{Tool Response Processing Approaches}
LLMs and tool use has been extensively studied, but most work focuses on the LLMs' ability to plan and formulate the tool calls correctly \citep{patil2023gorilla, schick2023toolformer, li2023api}. %Many of these use simple tools such as calculator and calendar with simple tool outputs. 
Benchmarks using real-world tools~\citep{qintoolllm, zhong2025complexfuncbench} acknowledge long and complex responses; \citet{qintoolllm} uses LLMs to prune parts of a long response and truncates it further to 1024 tokens.  \citet{song2023restgpt} has a code generation component to process API responses, however they do not evaluate that component in isolation or extensively on a large dataset. 

\citet{gu2024structexteval} evaluates LLMs on tasks on multiple structured formats including JSON, but their dataset is synthetically generated (failing to capture some properties of real-world tools), and they do not include evaluation of code generation.
% This work seems to evaluate generation of JSON:
Similarly, \citet{shorten2024structuredrag} evaluate LLMs' abilities to parse JSON responses, showing varying degrees of proficiency. However, their evaluation is restricted to one class of models and direct-prompting only. 
While table processing using LLMs fits within the structured output processing space~\citep{sui2024table, chen2024tablerag},
%~\citet{sui2024table} proposes a self-augmentation approach where the LLM is first prompted to generate some knowledge about its understanding of the table then include that in the subsequent prompt to extract specific information from the table; their results showed that table processing improves when using this simple yet generic approach. \citet{chen2024tablerag} adopts a schema and cell guided approach to table processing where they augment an LLM's prompt with relevant information (based on the question) about the table schema and the column-cell pairs. 
%However, unlike 
tables exhibit flatter structures than JSON structures which are highly nested, making them quite complex for the models to understand.

% Some existing agentic frameworks provide the ability to process JSON responses, including LangChain  \footnote{\url{https://python.langchain.com/docs/integrations/tools/json/}}, CrewAI \footnote{\url{https://github.com/alexfazio/crewAI-quickstart/blob/ main/crewai\_sequential\_JSONSearchTool\_quickstart.ipynb}} and SmythOS \footnote{\url{https://smythos.com/docs/agent-studio/components/advanced/json-filter/}}. They either rely on agent writers to define filtering criteria or use LLM-based processing. However, we could not find an analysis of how accurate these implementations are. 

\section{Conclusion}
Tool response processing is an important but non-trivial step in a tool-augmented LLM workflow. Understanding the characteristics of such processing and knowing when to use which approach is key to designing an effective system. This paper analyzes structured tool responses approaches and shows that even frontier models have room for improvement. Moreover, the performance varies significantly depending on the type of processing required and the prompting technique used. Our work can serve as guidance on how to approach this task and to drive future performance improvements.

\section*{Limitations}
Our work deals with a specific challenge in tool-augmented LLMs and hence we focus our experiments on that task. However, it would be important to evaluate the different tool response processing approaches in an end-to-end setting. Moreover, the processing approaches we considered are limited to answer and code generation, but there are other approaches such as RAG on the API response or an agentic approach to code generation. 

\section*{Ethics Statement}
We do not foresee any ethical concerns or opportunities to misuse this work. Furthermore, the data we used (and modified) to evaluate models does not contain any PII or harmful language. One possible consideration is whether the generate code may cause security concerns. However, there is nothing in our task that would prompt an LLM to attempt a security breach since the prompts all related to extracting answers from JSON responses. Whenever code is generated, there is always concern.   

\bibliography{custom}
\bibliographystyle{acl_natbib}

\appendix

% \section{Appendix}
% \label{sec:appendix}

\section{Dataset Examples}
Figures \ref{fig:extractive_templates}, \ref{fig:filtering_templates}, \ref{fig:aggregation_templates} show three examples from each category (extractive, filtering and aggregation) of QA templates from our dataset.

\begin{figure}[tb]
    % \centering
    \includegraphics[width=\linewidth]{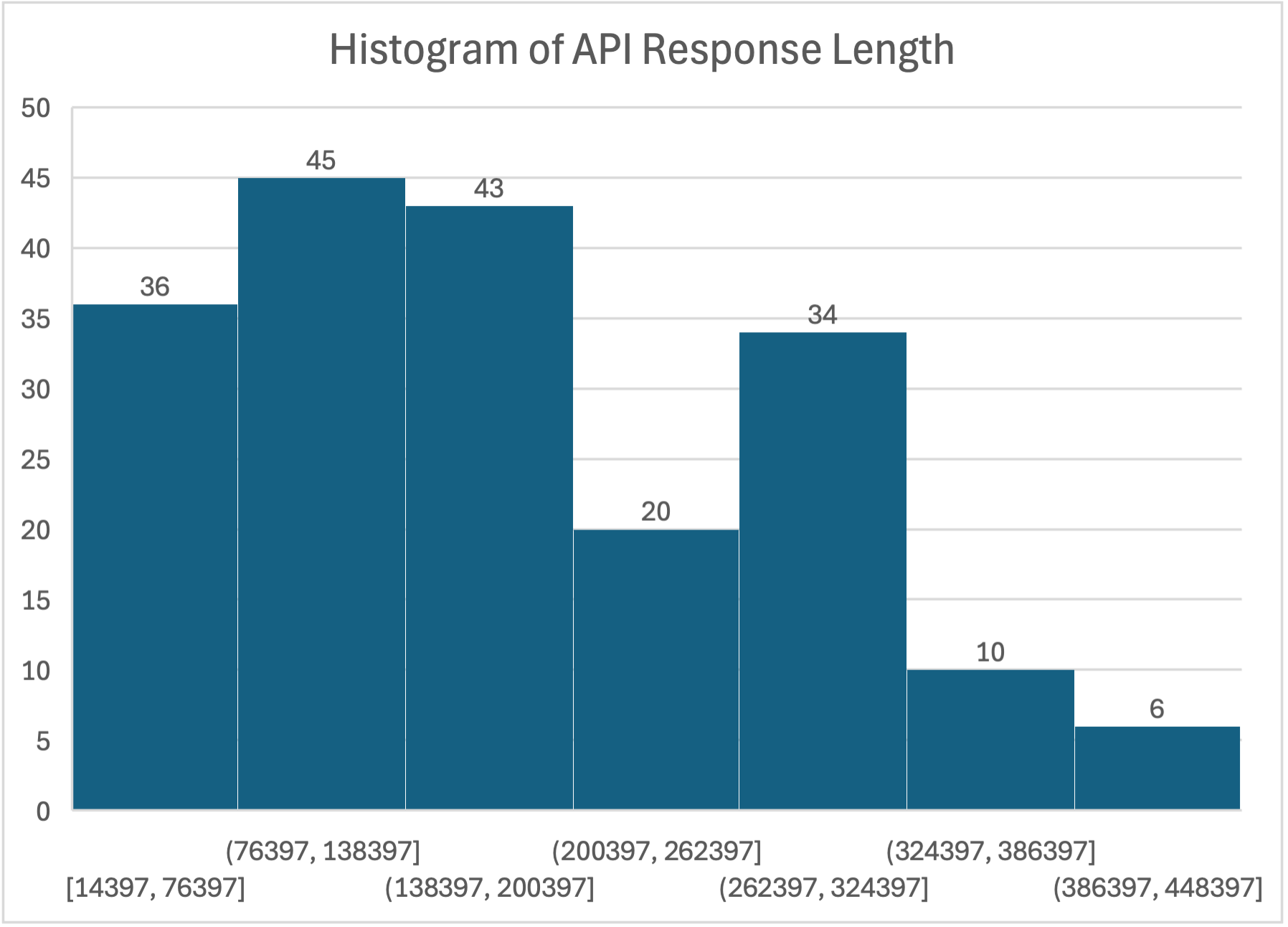}
    \caption{The histogram of API response lengths (number of characters) shows a large variance in response length for the same endpoint (Real Product - Shoes).}
    \label{fig:hist_product}
\end{figure}

Table \ref{tab:resp_stats} and Figure \ref{fig:hist_product} show that API responses can be exceptionally long. From the API responses we collected, about 20\% of the responses would not fit within a 65K context window and about 5\% would not fit in a 128K context window (assuming a tokenizer that maps about 4 characters to a token). However, given the processing approaches we wanted to compare, we filtered out such large responses from our dataset. In addition, we limit the list of models to those which support contexts with minimum 65K tokens. So all the API responses in our dataset fit within the context of the models we evaluated.

\begin{table}[thb]
\centering
% \scriptsize
\begin{tabular}{l c c}
\toprule
    & Real Product - Shoes & SEC-Filings   \\
    \midrule
Min & 14,397        & 38,642     \\
Max & 441,641       & 4,991,934   \\
Mean & 176,718.335   & 261,526    \\
Stdev & 102,880.236   & 782,602  \\
Median & 162,883	& 89,773 \\
\bottomrule
\end{tabular}
\caption{Statistics on the length (number of characters) of API responses of two endpoints; some endpoints may have significant variance in the length of the response. }
\label{tab:resp_stats}
\end{table}

\lstdefinestyle{listingstyle}{
    backgroundcolor=\color{backcolour},   
    commentstyle=\color{codegreen},
    keywordstyle=\color{magenta},
    numberstyle=\tiny\color{codegray},
    stringstyle=\color{codepurple},
    basicstyle=\fontsize{7}{7}\selectfont\ttfamily,
    breakatwhitespace=false,         
    breaklines=true,                 
    captionpos=b,                    
    keepspaces=true,                 
    numbers=left,                    
    numbersep=5pt,                  
    showspaces=false,                
    showstringspaces=false,
    showtabs=false,                
    tabsize=2,
    otherkeywords={Question, Template, Ground, Truth, Code, Example, Instance, Answer, Predicted, JSON, Snippet},
    morekeywords={Question, Template, Ground, Truth, Code, Example, Instance, Answer, Predicted, JSON, Snippet}
}
\lstset{style=listingstyle}

\begin{figure*}
\begin{lstlisting}[basicstyle=\fontsize{7}{7}\selectfont\ttfamily, label={lst:extractive_templates}]
------------------------------------------------------------------------------------------------------------
Extractive QA :
------------------------------------------------------------------------------------------------------------

Question Template: f"What is form type of the filing with accession number {acc_num}?" 
Ground Truth Code:
    def get_answer(self, acc_num: str, api_response: dict[Any, Any]) -> str:

        for forms in api_response["data"]["attributes"]["result"]:
            if forms["accessionNumber"] == acc_num:
                return str(forms["formType"])

        return "None"
Example Question Instance:What is form type of the filing with accession number 0001193125-23-072650? 
Example Answer Instance: ARS 

------------------------------------------------------------------------------------------------------------
                
Question Template: f"How many reviews does {hotel_name} have?" 
Ground Truth Code:    
    def get_answer(self, api_response: dict[Any, Any], hotel_name: str) -> str:
        for hotel in api_response["data"]["result"]:
            if hotel["hotel_name"].strip().lower() == hotel_name.strip().lower():
                return str(hotel["review_nr"])
        return "None"
Example Question Instance: How many reviews does "Terra Verde Vacation Rentals" have? 
Example Answer Instance: 19

------------------------------------------------------------------------------------------------------------

Question: f"What is the cleanliness rating of {vehicle_id}?"
Ground Truth Code:    
    def get_answer(self, api_response: dict[Any, Any], vehicle_id: str) -> str:
        for car in api_response["data"]["search_results"]:
            if car["vehicle_id"].strip().lower() == vehicle_id.strip().lower():
                return str(car["rating_info"]["cleanliness"])
        return "None"
Example Question Instance: What is the cleanliness rating of "370878816"?
Example Answer Instance: 8.5

------------------------------------------------------------------------------------------------------------

\end{lstlisting}
\caption{Three examples of extractive question templates and the code to extract the ground truth answer}
\label{fig:extractive_templates}
\end{figure*}

\begin{figure*}
\begin{lstlisting}[basicstyle=\fontsize{7}{7}\selectfont\ttfamily, label={lst:filtering_templates}]
------------------------------------------------------------------------------------------------------------
Filtering QA :
------------------------------------------------------------------------------------------------------------

Question Template: f"List the accession number of all the forms filed in year {year} which are of form type {form_type}. Output a comma separated list of accession numbers."
Ground Truth Code:
    def get_answer(
        self, api_response: dict[Any, Any], year: int, form_type: str
    ) -> str:

        access_num = []
        for filings in api_response["data"]["attributes"]["result"]:
            if "filingDate" in filings:
                date = datetime.strptime(filings["filingDate"], "%Y-%m-%dT%H:%M:%S")
                filing_year = date.year
                if year == filing_year:
                    if "formType" in filings and filings["formType"] == form_type:
                        access_num.append(filings["accessionNumber"])

        if len(access_num) != 0:
            return ",".join(access_num)
        else:
            return "None"
Example Question Instance: List the accession number of all the forms filed in year 2016 which are of form type CERTNYS. Output a comma separated list of accession numbers.
Example Answer Instance: 9999999997-16-023115

------------------------------------------------------------------------------------------------------------
            
Question Template: f"Which hotels have free parking? Output a comma separated list of hotel names."
Ground Truth Code: 
    def get_answer(self, api_response: dict[Any, Any]) -> str:
        hotels_list = []
        at_least_one = False
        for hotel in api_response["data"]["result"]:
            try:
                if hotel["has_free_parking"] == 1:
                    hotels_list.append(hotel["hotel_name"])
                    at_least_one = True
            except BaseException:
                continue
        if at_least_one:
            return ", ".join(hotels_list)
        else:
            return "None"
Example Question Instance: Which hotels have free parking? Output a comma separated list of hotel names. 
Example Answer Instance: SOLANA 7 Bed Rooms FAMILY VACATION VILLA-Private Pool and SPA, Casita mi Encanto a Cozy & Spacious House, 4-bedroom 3-Bath & pool, EXCELLENT 7 Bedroom EIR Vacation Villa - PRIVATE POOL, Contemporary Home w Private Pool& Spa, Terra Verde Vacation Rentals

------------------------------------------------------------------------------------------------------------

Question Template: f"Show me cars with prices in {currency}? Output a comma separated list of vehicle IDs." 
Ground Truth Code:
    def get_answer(self, api_response: dict[Any, Any], currency: str) -> str:
        vehicle_list = []
        at_least_one = False
        for car in api_response["data"]["search_results"]:
            try:
                if car["pricing_info"]["base_currency"] == currency:
                    vehicle_list.append(car["vehicle_id"])
                    at_least_one = True
            except BaseException:
                continue
        if at_least_one:
            return ", ".join(vehicle_list)
        else:
            return "None"
Example Question Instance: Show me cars with prices in INR? Output a comma separated list of vehicle IDs. 
Example Answer Instance: 370878721, 370878796, 370878761, 370878771, 370878816

------------------------------------------------------------------------------------------------------------
\end{lstlisting}
\caption{Three examples of filtering question templates and the code to extract the ground truth answer}
\label{fig:filtering_templates}
\end{figure*}

\begin{figure*}
\begin{lstlisting}[basicstyle=\fontsize{7}{7}\selectfont\ttfamily, label={lst:aggregation_templates}]
------------------------------------------------------------------------------------------------------------
Aggregation QA:
------------------------------------------------------------------------------------------------------------

Question Template: f"How many SEC filings are done in year {year}?"
Ground Truth Code:
    def get_answer(self, api_response: dict[Any, Any], year: int) -> str:

        reports = 0
        for filings in api_response["data"]["attributes"]["result"]:
            date = datetime.strptime(filings["filingDate"], "%Y-%m-%dT%H:%M:%S")
            filing_year = date.year
            if filing_year == year:
                reports += 1

        return str(reports)
Example Question Instance: How many SEC filings are done in year 2022?
Example Answer Instance: 32

------------------------------------------------------------------------------------------------------------

Question Template: f"What is the average price reported for these hotels?"
Ground Truth Code:
    def get_answer(self, api_response: dict[Any, Any]) -> str:
        price_list = []
        for hotel in api_response["data"]["result"]:
            price_list.append(hotel["min_total_price"])
        return str(sum(price_list) / len(price_list))
Example Question Instance: What is the average price reported for these hotels?
Example Answer Instance: 464.746

------------------------------------------------------------------------------------------------------------

Question Template: f"What is the cheapest base price available?"
Ground Truth Code:
    def get_answer(self, api_response: dict[Any, Any]) -> str:
        car_price = []
        for car in api_response["data"]["search_results"]:
            car_price.append(car["pricing_info"]["base_price"])
        return str(min(car_price))
Example Question Instance: What is the cheapest base price available?
Example Answer Instance: 23288.56

------------------------------------------------------------------------------------------------------------
\end{lstlisting}
\caption{Three examples of aggregation question templates and the code to extract the ground truth answer}
\label{fig:aggregation_templates}
\end{figure*}

\section{Model Details}
\label{appendix:model_details}

Table \ref{tab:models} lists all the models used in this study, their sizes and the pointers to the exact model.
        
\begin{table*}[tbh]
    \centering
    \scriptsize
    \begin{tabular}{lrl}
    \toprule
        Model & \# Active Parameters & Model Pointer \\
        \midrule
        Llama-3.1-8B-Instruct & 8B & https://huggingface.co/meta-llama/Llama-3.1-8B-Instruct \\
        granite-3.1-8b-instruct & 8B & https://huggingface.co/ibm-granite/granite-3.1-8b-instruct \\        
        Granite-3.3-8b-instruct & 8B & https://huggingface.co/ibm-granite/granite-3.3-8b-instruct \\
        Llama-4-maverick-17b-128e-instruct-fp8 & 17B & https://huggingface.co/meta-llama/Llama-4-Maverick-17B-128E-Instruct-FP8\\
        GPT-OSS-20B & 20B & https://huggingface.co/openai/gpt-oss-20b \\
        Devstral-Small-2507 & 24B & https://huggingface.co/mistralai/Devstral-Small-2507 \\
        Deepseek-V3 & 37B & https://huggingface.co/deepseek-ai/DeepSeek-V3 \\
        Mixtral-8x22B-instruct-v0.1 & 39.1B & https://huggingface.co/mistralai/Mixtral-8x22B-Instruct-v0.1 \\
        Llama-3.1-70B-Instruct & 70B & https://huggingface.co/meta-llama/Llama-3.1-70B-Instruct \\
        Llama-3-3-70b-instruct & 70B & https://huggingface.co/meta-llama/Llama-3.3-70B-Instruct \\
        DeepSeek-R1-Distill-Llama-70B & 70B & https://huggingface.co/deepseek-ai/DeepSeek-R1-Distill-Llama-70B \\
        GPT-OSS-120b & 120B & https://huggingface.co/openai/gpt-oss-120b \\
        Mistral-large & 123B & https://huggingface.co/mistralai/Mistral-Large-Instruct-2411  \\
        Qwen3-235B-A22B-Instruct-2507 &  235B & https://huggingface.co/Qwen/Qwen3-235B-A22B-Instruct-2507 \\
        Llama-3-405B-Instruct & 405B & https://huggingface.co/meta-llama/Llama-3.1-405B-Instruct \\
        Qwen3-Coder-480B-A35B-Instruct-FP8 & 480B & https://huggingface.co/Qwen/Qwen3-Coder-480B-A35B-Instruct-FP8 \\
        GPT-4o-2024-11-20 & Unknown & https://openai.com/index/hello-gpt-4o/ \\
        Claude-4-sonnet & Unknown & https://www.anthropic.com/claude/sonnet \\
     \bottomrule           
    \end{tabular}
    \caption{A list of all the models used in our study (sorted by size of active parameters, if known).}
    \label{tab:models}
\end{table*}

\section{Tool Response Processing Approaches}

\subsection{Reduced Response - Heuristic Approach}
In the reduced response settings in RQ5, we reduced the length of the response using a heuristic approach. Given our observation that LLMs still benefit from the response because it includes information missing from the schema (such as sample values), we implement an algorithm that keeps one sample of each unique key. Essentially, we eliminate all lists (by keeping only one element). This allows us to present the structure of the JSON and the dependencies between keys while reducing the length. 

\subsection{Schema}
We generate the schema from an API response by using an existing Python library that can do so\footnote{https://transform.tools/json-to-json-schema}. 
Figure \ref{fig:schema_sample} shows one example of a scheme generated for the SEC-Filings endpoint.

\begin{figure*}
\begin{lstlisting}[basicstyle=\fontsize{7}{7}\selectfont\ttfamily, label={lst:sample_schema}]
------------------------------------------------------------------------------------------------------------
Sample Schema
------------------------------------------------------------------------------------------------------------
{
        "$schema": "http://json-schema.org/draft-07/schema#",
        "title": "Response schema for the SEC filings endpoint.",
        "type": "object",
        "properties": {
            "meta": {
            "type": "object",
            "properties": {
                "copyright": { "type":  "string" },
                "terms": { "type": "array", "items": { "type": "string"  }  }
            },
            "required": [  "copyright", "terms" ]
            },
            "data": {
            "type": "object",
            "properties": {
                "attributes": {
                "type": "object",
                "properties": {
                    "status": {
                    "type": "number"
                    },
                    "company": {
                    "type": "object",
                    "properties": {
                        "name": {
                        "type": "string"
                        },
                        "cik": {
                        "type": "string"
                        },
                        "ticker": {
                        "type": "string"
                        }
                    },
                    "required": [ "name",  "cik", "ticker" ]
                    },
                    "count": {
                    "type": "number"
                    },
                    "result": {
                    "type": "array",
                    "items": {
                        "type": "object",
                        "properties": {
                        "name": {
                            "type": "string"
                        },
                        "accessionNumber": {
                            "type": "string"
                        },
                        "filingDate": {
                            "type": "string"
                        },
                        "formType": {
                            "type": "string"
                        },
                        "url": {
                            "type": "string"
                        },
                        "period": {
                            "type": "string"
                        }
                        },
                        "required": [ "name", "accessionNumber", "filingDate", "formType", "url"  ]
                    }
                    }
                },
                "required": [  "status", "company", "count",  "result" ]
                },
                "type": { "type": "string" },
                "id": { "type": "string" }
            },
            "required": [  "attributes", "type", "id" ]
            },
            "status": { "type": "boolean" }
        },
        "required": [ "meta", "data", "status" ]
    }
------------------------------------------------------------------------------------------------------------

\end{lstlisting}
\caption{Example of an API response's schema}
\label{fig:schema_sample}
\end{figure*}

\subsection{Prompts}
\label{sec:prompts}

Figures~\ref{fig:prompt_direct_prompting}, \ref{fig:prompt_direct_prompting_schema} \ref{fig:prompt_code_gen_schema}, \ref{fig:prompt_code_gen}, 
\ref{fig:prompt_code_gen_no_resp}, \ref{fig:prompt_code_gen_compact_resp}, \ref{fig:prompt_cot_direct_prompting_schema}, and \ref{fig:prompt_cot_code_gen_schema} provide prompts for the settings mentioned in Section \ref{sec:tool_response_processing}.
Figure~\ref{fig:prompt_llm_judge} shows the prompt used for the \textit{LLM as a judge} metric. Figure~\ref{fig:prompt_classifier} shows the prompt used for the query classifier.

\begin{figure*}
\begin{lstlisting}[basicstyle=\fontsize{7}{7}\selectfont\ttfamily, label={lst:code_gen_1}]
------------------------------------------------------------------------------------------------------------
Zero-shot prompt for model output type = answer, without a schema: 
------------------------------------------------------------------------------------------------------------

You are given a response from an API call (in JSON format).
Answer the question based on the information provided in the API response.

```json
{api_response}
```

Question: {question}

Only respond with the answer. Do not include any other text or json in the response.
Do not rephrase the answer or write it in complete sentence, return exactly as is from the JSON.

Answer:
------------------------------------------------------------------------------------------------------------
\end{lstlisting}
\caption{Prompt for model output type = answer, without a schema}
\label{fig:prompt_direct_prompting}
\end{figure*}

\begin{figure*}
\begin{lstlisting}[basicstyle=\fontsize{7}{7}\selectfont\ttfamily, label={lst:code_gen_2}]
------------------------------------------------------------------------------------------------------------
Zero-shot prompt for model output type = answer, with a schema: 
------------------------------------------------------------------------------------------------------------

You will be given a JSON object as data which is a response from a REST API containing information returned from the API call.
You are given a response from an API call (in JSON format). 
Answer the question based on the information provided in the API response.

```json
{api_response}
```

Question: {question}

Only respond with the answer. Do not include any other text or json in the response.
The JSON schema of the object given as data is as follows: {json_schema}

Answer:
------------------------------------------------------------------------------------------------------------
\end{lstlisting}
\caption{Prompt for model output type = answer, with a schema}
\label{fig:prompt_direct_prompting_schema}
\end{figure*}

\begin{figure*}
\begin{lstlisting}[basicstyle=\fontsize{7}{7}\selectfont\ttfamily, label={lst:code_gen_3}]
------------------------------------------------------------------------------------------------------------
Zero-shot prompt for model output type = answer, without a schema: 
------------------------------------------------------------------------------------------------------------

You will be given a JSON object as data which is a response from a REST API containing information returned from the API call.
Your task is to extract and return <<task_prefix>>

Write a Python function that:
    Starts the function with "def ".
    Takes only the entire api response as input and doesn't have any other input.
    Identifies the structure of the input data, ensuring it checks for relevant keys and data types.
    When comparing strings, it should always convert both sides of the comparison to lowercase.
    Processes the provided data.
    Iterates through the data to extract relevant information.
    Cleans numeric strings by removing non-numeric characters before converting them to integers.
    Performs proper checks to ensure a key exists and is not None before querying its value.
    Returns only the requested data as a string and no other extra information or words.
    Do not add any extra keys or terms to the output.

Final Check:
    The function must be formatted in Python markdown for direct execution.
    No explanations, comments, or additional text should be included.
    Do not include any example usage.

data = <<json_obj>>

Python Function:

------------------------------------------------------------------------------------------------------------
\end{lstlisting}
\caption{Prompt for model output type = answer, without a schema}
\label{fig:prompt_code_gen}
\end{figure*}

\begin{figure*}
\begin{lstlisting}[basicstyle=\fontsize{7}{7}\selectfont\ttfamily, label={lst:code_gen_schema_1}]
------------------------------------------------------------------------------------------------------------
Zero-shot prompt for model output type = code, with a schema: 
------------------------------------------------------------------------------------------------------------

You will be given a JSON object as data which is a response from a REST API containing information returned from the API call.
You will be given a JSON schema of the response from the REST API returned from the API call.
Your task is to extract and return information from the JSON object which follows the JSON schema and answers the user query: <<task_prefix>>

You need to write a Python function that:
    Starts the function with "def ".
    Takes only the entire api response as input and doesn't have any other input.
    Identifies the structure of the input data, ensuring it checks for relevant keys and data types.
    When comparing strings, it should always convert both sides of the comparison to lowercase.
    Processes the provided data.
    Iterates through the data to extract relevant information.
    Cleans numeric strings by removing non-numeric characters before converting them to integers.
    Performs proper checks to ensure a key exists and is not None before querying its value.
    Returns only the requested data as a string and no other extra information or words.
    Do not add any extra keys or terms to the output.

Final Check:
    The function must be formatted in Python markdown for direct execution.
    No explanations, comments, or additional text should be included.
    Do not include any example usage data.

The JSON schema of the object given as data is as follows: <<json_schema>>

data = <<json_obj>>

Python Function:
------------------------------------------------------------------------------------------------------------
\end{lstlisting}
\caption{Prompt for model output type = code, with a schema}
\label{fig:prompt_code_gen_schema}
\end{figure*}

\begin{figure*}
\begin{lstlisting}[basicstyle=\fontsize{7}{7}\selectfont\ttfamily, label={lst:code_gen_4}]
------------------------------------------------------------------------------------------------------------
Zero-shot prompt for model output type = code, with a schema but without a response: 
------------------------------------------------------------------------------------------------------------
Your task is to work with an already-loaded JSON object as a dictionary from a REST API response. 
Using the provided JSON schema, you need to extract and return information that directly answers the user's query:
<<task_prefix>>

You need to write a Python function that:
    Starts the function with "def ".
    Takes only the entire api response as input and doesn't have any other input.
    Identifies the structure of the input data, ensuring it checks for relevant keys and data types.
    When comparing strings, it should always convert both sides of the comparison to lowercase.
    Processes the provided data.
    Iterates through the data to extract relevant information.
    Cleans numeric strings by removing non-numeric characters before converting them to integers.
    Performs proper checks to ensure a key exists and is not None before querying its value.
    Returns only the requested data as a string and no other extra information or words.
    Do not add any extra keys or terms to the output.

Final Check:
    The function must be formatted in Python markdown for direct execution.
    No explanations, comments, or additional text should be included.
    Do not include any example usage data.

The JSON schema of the object given as data is as follows: <<json_schema>>

Python Function:
------------------------------------------------------------------------------------------------------------
\end{lstlisting}
\caption{Prompt for model output type = code, with a schema but without a response}
\label{fig:prompt_code_gen_no_resp}
\end{figure*}

\begin{figure*}
\begin{lstlisting}[basicstyle=\fontsize{7}{7}\selectfont\ttfamily, label={lst:code_gen_5}]
------------------------------------------------------------------------------------------------------------
Zero-shot prompt for model output type = code, with a schema but with a reduced response: 
------------------------------------------------------------------------------------------------------------
You will be given a JSON object as data which is an example or compact version of the response from a REST API containing information returned from the API call.
You will be given a JSON schema of the response from the REST API returned from the API call.
Your task is to take reference and return information from the JSON object which follows the JSON schema and answers the user query: <<task_prefix>>

You need to write a Python function that:
    Starts the function with "def ".
    Takes only the entire api response as input and doesn't have any other input.
    Identifies the structure of the input data, ensuring it checks for relevant keys and data types.
    When comparing strings, it should always convert both sides of the comparison to lowercase. This is mandatory.
    Processes the provided data.
    Iterates through the data to extract relevant information.
    Cleans numeric strings by removing non-numeric characters before converting them to integers.
    Performs proper checks to ensure a key exists and is not None before querying its value.
    Returns only the requested data as a string and no other extra information or words.
    The data to be returned from the function should be in string.
    Do not add any extra keys or terms to the output.

Final Check:
    The function must be formatted in Python markdown for direct execution.
    There should be ```python in the beginning and the ending should be ```.
    No explanations, comments, or additional text should be included.
    Do not include any example usage data.

The JSON schema of the object given as data is as follows: <<json_schema>>

data = <<json_obj>>

Python Function:
------------------------------------------------------------------------------------------------------------
\end{lstlisting}
\caption{Prompt for model output type = code, with a schema but with a reduced response}
\label{fig:prompt_code_gen_compact_resp}
\end{figure*}

\begin{figure*}
\begin{lstlisting}[basicstyle=\fontsize{7}{7}\selectfont\ttfamily, label={lst:code_gen_6}]
------------------------------------------------------------------------------------------------------------
Zero-shot prompt for CoT, model output type = answer, with a schema: 
------------------------------------------------------------------------------------------------------------

You will be given a JSON object as data which is a response from a REST API containing information returned from the API call.
You are given a response from an API call (in JSON format).
Answer the question based on the information provided in the API response.

```json
{api_response}
```

Question: {question}

Think step-by-step about how the data relates to the question being asked.
Analyze the structure and contents of the JSON response.
Identify which fields are relevant to the question.
Then, reason through the data logically to derive the answer.

Finally, provide the answer accurately.

Only respond with the answer. Do not include any other text or json in the response.

The JSON schema of the object given as data is as follows: <<json_schema>>

Answer:
        
------------------------------------------------------------------------------------------------------------

\end{lstlisting}
\caption{Prompt for CoT, model output type = answer, with a schema}
\label{fig:prompt_cot_direct_prompting_schema}
\end{figure*}
\begin{figure*}

\begin{lstlisting}[basicstyle=\fontsize{7}{7}\selectfont\ttfamily, label={lst:code_gen_7}]
------------------------------------------------------------------------------------------------------------
Zero-shot prompt for CoT, model output type = code, with a schema: 
------------------------------------------------------------------------------------------------------------

You will be given a JSON object as data which is a response from a REST API containing information returned from the API call.
You will be given a JSON schema of the response from the REST API returned from the API call.
Your task is to extract and return information from the JSON object which follows the JSON schema and answers the user query: <<task_prefix>>


Think step-by-step:
    - First, analyze the structure of the JSON data using the schema.
    - Identify which fields are relevant to the task and understand their data types.
    - Consider edge cases like missing keys, null values, or unexpected types.
    - When comparing strings, normalize them to lowercase.
    - When handling numeric strings, remove non-numeric characters before converting to integers.
    - Use logical iteration and condition checks to extract the required information.
    - Ensure that only the expected value is returned, formatted strictly as a string.

You need to write a Python function that:
    Starts the function with "def ".
    Takes only the entire api response as input and doesn't have any other input.
    Identifies the structure of the input data, ensuring it checks for relevant keys and data types.
    Processes the provided data.
    Iterates through the data to extract relevant information.
    Cleans numeric strings by removing non-numeric characters before converting them to integers.
    Performs proper checks to ensure a key exists and is not None before querying its value.
    Returns only the requested data as a string and no other extra information or words.
    Do not add any extra keys or terms to the output.

Final Check:
    The function must be formatted in Python markdown for direct execution.
    No explanations, comments, or additional text should be included.
    Do not include any example usage data.

The JSON schema of the object given as data is as follows: <<json_schema>>

data = <<json_obj>>

Python Function:

------------------------------------------------------------------------------------------------------------
\end{lstlisting}
\caption{Prompt for CoT, model output type = code, with a schema}
\label{fig:prompt_cot_code_gen_schema}
\end{figure*}

\begin{figure*}
\begin{lstlisting}[basicstyle=\fontsize{7}{7}\selectfont\ttfamily, label={lst:llm_as_a_judge_1}]
------------------------------------------------------------------------------------------------------------
Prompt for LLM as a judge: 
------------------------------------------------------------------------------------------------------------

You are given two inputs, one is the ground truth and one is predicted answer.
These are two strings. Your task is to output either "True" or "False" indicating whether the
predicted answer matches the ground truth.
Please follow the instructions listed below:
1. As these are free text outputs, the match does not need to be exact.
2. Please allow for paraphrases or verbalizations.
3. Please output only "True" or "False".

ground_truth:{gold_answer}
predicted_answer:{predicted_answer}

Does the predicted_answer match the ground_truth?: 
------------------------------------------------------------------------------------------------------------
\end{lstlisting}
\caption{Prompt for the LLM as a judge metric}
\label{fig:prompt_llm_judge}
\end{figure*}

\begin{figure*}
\begin{lstlisting}[basicstyle=\fontsize{7}{7}\selectfont\ttfamily, label={lst:llm_as_a_judge_2}]
------------------------------------------------------------------------------------------------------------
Prompt for classifying queries: 
------------------------------------------------------------------------------------------------------------

You will be given a natural language query and are tasked with classifying it into one of the following three categories:
- Extractive: returns a value in the JSON given a key
- Filtering: returns multiple entries corresponding to filtering criteria
- Aggregation: combines multiple entries by performing an aggregation operation

Return your final answer after "Final answer, category: "

Does the following natural language query belong to the "Extractive", "Filtering" or "Aggregation" category?

Natural language query: {nl_query}

Schema: {schema}

Final answer, category: 

------------------------------------------------------------------------------------------------------------
\end{lstlisting}
\caption{Prompt for classifying queries int extractive, filtering or aggregation}
\label{fig:prompt_classifier}
\end{figure*}

\section{Detailed Results}
\subsection{RQ3 and RQ4}
Figure \ref{fig:rq3_4_exact_match} illustrates the exact match results of four prompting strategies across 15 models. Since the values may be difficult to read, Table \ref{tab:rq3_4_exact_match_nbs} reports the exact numbers used to generate Figure \ref{fig:rq3_4_exact_match}.

\begin{table*}[ht]
\centering
\scriptsize
\begin{tabular}{lllllcccc}
\toprule
Model                              & Model Output Type & Reasoning & Schema & Response & Total & Extractive & Filtering & Aggregation \\
\midrule
Granite-3.3-8b-instruct            & Code              & None      & None   & Full     & 0.325                     & 0.426                                & 0.183                               & 0.318                                 \\
Llama-4-maverick-17b-128e-instruct & Code              & None      & None   & Full     & 0.720                     & 0.824                                & 0.574                               & 0.710                                 \\
GPT-oss-20b                        & Code              & None      & None   & Full     & 0.615                     & 0.679                                & 0.429                               & 0.725                                 \\
Devstral-Small-2507                & Code              & None      & None   & Full     & 0.672                     & 0.719                                & 0.558                               & 0.725                                 \\
DeepSeek-V3                        & Code              & None      & None   & Full     & 0.762                     & 0.848                                & 0.609                               & 0.793                                 \\
Mixtral-8x22B-instruct-v0.1        & Code              & None      & None   & Full     & 0.315                     & 0.319                                & 0.231                               & 0.410                                 \\
Llama-3-3-70b-instruct             & Code              & None      & None   & Full     & 0.666                     & 0.781                                & 0.497                               & 0.667                                 \\
DeepSeek-R1-Distill-Llama-70B      & Code              & None      & None   & Full     & 0.676                     & 0.807                                & 0.513                               & 0.639                                 \\
GPT-oss-120b                       & Code              & None      & None   & Full     & 0.754                     & 0.783                                & 0.604                               & 0.886                                 \\
Mistral-large                      & Code              & None      & None   & Full     & 0.653                     & 0.671                                & 0.627                               & 0.654                                 \\
Qwen3-235B-A22B-Instruct-2507      & Code              & None      & None   & Full     & 0.704                     & 0.734                                & 0.602                               & 0.775                                 \\
Llama-3-405B-Instruct              & Code              & None      & None   & Full     & 0.687                     & 0.784                                & 0.518                               & 0.719                                 \\
Qwen3-Coder-480B-A35B-Instruct-FP8 & Code              & None      & None   & Full     & 0.765                     & 0.809                                & 0.640                               & 0.840                                 \\
GPT-4o                             & Code              & None      & None   & Full     & 0.703                     & 0.726                                & 0.637                               & 0.741                                 \\
Claude-4-Sonnet                    & Code              & None      & None   & Full     & 0.624                     & 0.653                                & 0.500                               & 0.722                                 \\
Granite-3.3-8b-instruct            & Code              & None      & Full   & Full     & 0.451                     & 0.607                                & 0.266                               & 0.398                                 \\
Llama-4-maverick-17b-128e-instruct & Code              & None      & Full   & Full     & 0.763                     & 0.824                                & 0.596                               & 0.855                                 \\
GPT-oss-20b                        & Code              & None      & Full   & Full     & 0.723                     & 0.824                                & 0.472                               & 0.846                                 \\
Devstral-Small-2507                & Code              & None      & Full   & Full     & 0.671                     & 0.778                                & 0.462                               & 0.735                                 \\
DeepSeek-V3                        & Code              & None      & Full   & Full     & 0.781                     & 0.767                                & 0.746                               & 0.849                                 \\
Mixtral-8x22B-instruct-v0.1        & Code              & None      & Full   & Full     & 0.414                     & 0.424                                & 0.345                               & 0.478                                 \\
Llama-3-3-70b-instruct             & Code              & None      & Full   & Full     & 0.713                     & 0.803                                & 0.541                               & 0.762                                 \\
DeepSeek-R1-Distill-Llama-70B      & Code              & None      & Full   & Full     & 0.679                     & 0.819                                & 0.528                               & 0.611                                 \\
GPT-oss-120b                       & Code              & None      & Full   & Full     & 0.832                     & 0.864                                & 0.675                               & 0.966                                 \\
Mistral-large                      & Code              & None      & Full   & Full     & 0.734                     & 0.781                                & 0.759                               & 0.620                                 \\
Qwen3-235B-A22B-Instruct-2507      & Code              & None      & Full   & Full     & 0.761                     & 0.747                                & 0.647                               & 0.926                                 \\
Llama-3-405B-Instruct              & Code              & None      & Full   & Full     & 0.701                     & 0.850                                & 0.449                               & 0.741                                 \\
Qwen3-Coder-480B-A35B-Instruct-FP8 & Code              & None      & Full   & Full     & 0.801                     & 0.845                                & 0.680                               & 0.870                                 \\
GPT-4o                             & Code              & None      & Full   & Full     & 0.751                     & 0.797                                & 0.607                               & 0.846                                 \\
Claude-4-Sonnet                    & Code              & None      & Full   & Full     & 0.743                     & 0.788                                & 0.642                               & 0.784                                 \\
Granite-3.3-8b-instruct            & Answer            & None      & None   & Full     & 0.519                     & 0.857                                & 0.157                               & 0.355                                 \\
Llama-4-maverick-17b-128e-instruct & Answer            & None      & None   & Full     & 0.643                     & 0.917                                & 0.353                               & 0.503                                 \\
GPT-oss-20b                        & Answer            & None      & None   & Full     & 0.260                     & 0.271                                & 0.000                               & 0.556                                 \\
Devstral-Small-2507                & Answer            & None      & None   & Full     & 0.634                     & 0.938                                & 0.287                               & 0.512                                 \\
DeepSeek-V3                        & Answer            & None      & None   & Full     & 0.649                     & 0.872                                & 0.416                               & 0.534                                 \\
Mixtral-8x22B-instruct-v0.1        & Answer            & None      & None   & Full     & 0.563                     & 0.819                                & 0.307                               & 0.417                                 \\
Llama-3-3-70b-instruct             & Answer            & None      & None   & Full     & 0.368                     & 0.490                                & 0.030                               & 0.562                                 \\
DeepSeek-R1-Distill-Llama-70B      & Answer            & None      & None   & Full     & 0.312                     & 0.376                                & 0.008                               & 0.568                                 \\
GPT-oss-120b                       & Answer            & None      & None   & Full     & 0.284                     & 0.309                                & 0.000                               & 0.583                                 \\
Mistral-large                      & Answer            & None      & None   & Full     & 0.608                     & 0.869                                & 0.345                               & 0.460                                 \\
Qwen3-235B-A22B-Instruct-2507      & Answer            & None      & None   & Full     & 0.399                     & 0.441                                & 0.170                               & 0.602                                 \\
Llama-3-405B-Instruct              & Answer            & None      & None   & Full     & 0.670                     & 0.926                                & 0.411                               & 0.528                                 \\
Qwen3-Coder-480B-A35B-Instruct-FP8 & Answer            & None      & None   & Full     & 0.654                     & 0.903                                & 0.322                               & 0.611                                 \\
GPT-4o                             & Answer            & None      & None   & Full     & 0.658                     & 0.914                                & 0.444                               & 0.460                                 \\
Claude-4-Sonnet                    & Answer            & None      & None   & Full     & 0.700                     & 0.948                                & 0.419                               & 0.596                                 \\
Granite-3.3-8b-instruct            & Answer            & None      & Full   & Full     & 0.383                     & 0.540                                & 0.195                               & 0.330                                 \\
Llama-4-maverick-17b-128e-instruct & Answer            & None      & Full   & Full     & 0.648                     & 0.910                                & 0.393                               & 0.488                                 \\
GPT-oss-20b                        & Answer            & None      & Full   & Full     & 0.273                     & 0.298                                & 0.000                               & 0.562                                 \\
Devstral-Small-2507                & Answer            & None      & Full   & Full     & 0.640                     & 0.941                                & 0.305                               & 0.509                                 \\
DeepSeek-V3                        & Answer            & None      & Full   & Full     & 0.666                     & 0.936                                & 0.398                               & 0.506                                 \\
Mixtral-8x22B-instruct-v0.1        & Answer            & None      & Full   & Full     & 0.543                     & 0.783                                & 0.287                               & 0.426                                 \\
Llama-3-3-70b-instruct             & Answer            & None      & Full   & Full     & 0.297                     & 0.369                                & 0.020                               & 0.506                                 \\
DeepSeek-R1-Distill-Llama-70B      & Answer            & None      & Full   & Full     & 0.331                     & 0.398                                & 0.020                               & 0.586                                 \\
GPT-oss-120b                       & Answer            & None      & Full   & Full     & 0.280                     & 0.316                                & 0.000                               & 0.559                                 \\
Mistral-large                      & Answer            & None      & Full   & Full     & 0.592                     & 0.848                                & 0.322                               & 0.460                                 \\
Qwen3-235B-A22B-Instruct-2507      & Answer            & None      & Full   & Full     & 0.295                     & 0.333                                & 0.008                               & 0.577                                 \\
Llama-3-405B-Instruct              & Answer            & None      & Full   & Full     & 0.653                     & 0.921                                & 0.355                               & 0.534                                 \\
Qwen3-Coder-480B-A35B-Instruct-FP8 & Answer            & None      & Full   & Full     & 0.644                     & 0.821                                & 0.437                               & 0.580                                 \\
GPT-4o                             & Answer            & None      & Full   & Full     & 0.644                     & 0.905                                & 0.419                               & 0.451                                 \\
Claude-4-Sonnet                    & Answer            & None      & Full   & Full     & 0.704                     & 0.957                                & 0.447                               & 0.565    \\                            
\bottomrule
\end{tabular}
\caption{Exact match accuracy results for RQ3 and RQ4 across 15 models indicate that code generation often outperforms answer generation on filtering and aggregation across models; including the schema almost always boosts results for all settings. Prompts did not include CoT reasoning and used the full API responses.}
\label{tab:rq3_4_exact_match_nbs}
\end{table*}

\subsection{RQ6}
Filtering and aggregation generally do better with reasoning prompts whether generating the answer or code (12 and 14 models, respectively). Extractive does better without reasoning instructions  (14 out of 15 models). When comparing answer vs. code generation with CoT, the former beats the latter on filtering questions for 8 out of 15 models compared to 6 out of 15 for non-reasoning answer vs. code generation comparison.; 10 out of 15 on aggregation questions compared to only 5.

% \begin{figure}[tb]
%     \centering
%     \includegraphics[width=1\linewidth]
%     {figures/rq6_contains_x3.pdf}
%     \caption{Reasoning-based prompting results}
%     \label{fig:rq6_contains}
% \end{figure}

\begin{figure*}[tb]
    \centering
    \includegraphics[width=1\linewidth]
    {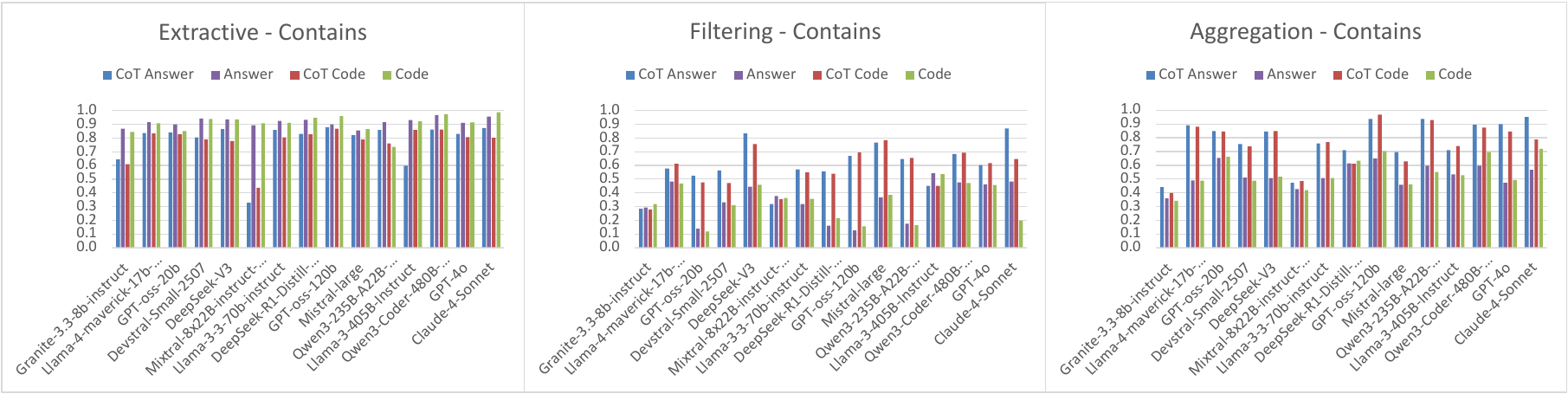}
    \caption{Detailed reasoning-based prompting (i.e., with and without CoT) results for the 15 models indicate that filtering and aggregation question benefit most from CoT prompts.}
    \label{fig:rq6_contains}
\end{figure*}

\section{Impact of API Structure (RQ8)}
% setup
Our dataset contained six endpoints from different domains and each QA template required processing the JSON to varying depths. In the results below, we attempt to identify correlations between the endpoints and nesting of the JSON.

\begin{figure*}[tb]
    \centering
    \includegraphics[width=\linewidth]{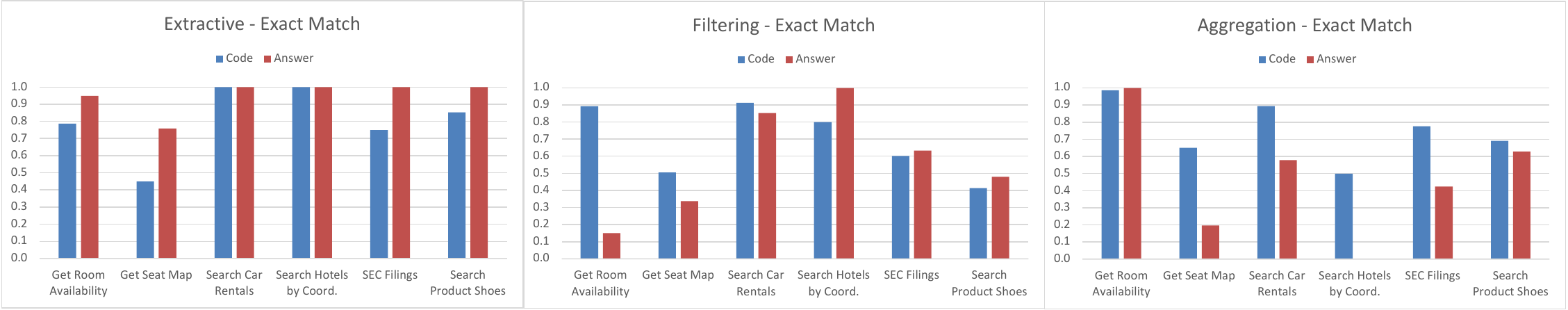}
    \caption{The exact match accuracy per endpoint for Claude-4-Sonnet shows that there is no clear correlation between endpoints and performance of either prompting approach. }
    \label{fig:rq8_endpoint}
\end{figure*}

% \subsubsection{Per Endpoint} 
Figure \ref{fig:rq8_endpoint} plots the accuracy per endpoint for Claude-4-Sonnet. We do not observe any correlation between endpoints and which approach is more suitable. 

% \begin{figure}[h!]
%     \centering
%     \includegraphics[width=\linewidth]{figures/rq8_per_endpoint.pdf}
%     \caption{Exact Match Accuracy per Endpoint for Claude-4-Sonnet Exact Match}
%     \label{fig:rq8_endpoint}
% \end{figure}

\begin{figure}[tb]
    \centering
    \includegraphics[width=\linewidth]{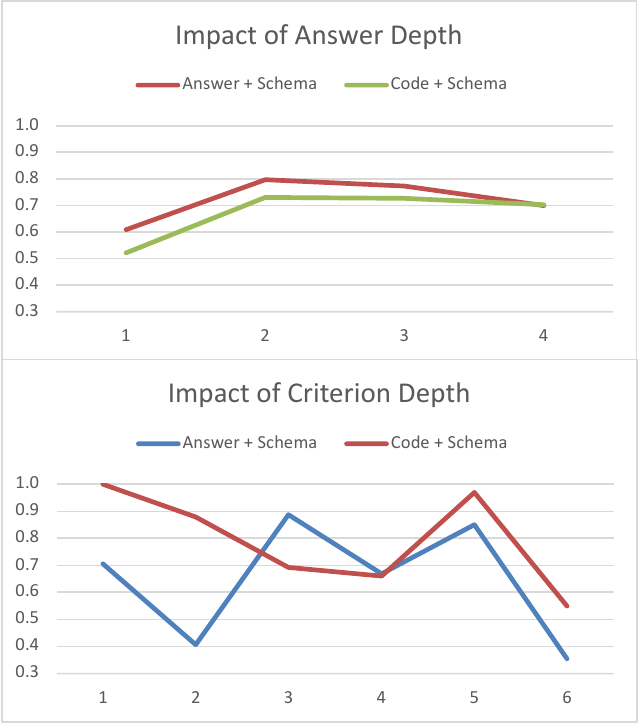}
    \caption{The exact match accuracy per depth of answer or filtering criterion for Claude-4-Sonnet. No clear correlation between accuracy and depth. }
    \label{fig:rq8_depth}
\end{figure}

% \subsubsection{Nesting}
Figure \ref{fig:rq8_depth} groups the QA pairs by the depth of the answer or the filtering criterion. We do not observe any clear correlation between nesting and performance. Thus, we cannot conclude whether answer generation was better than code generation beyond some threshold of depth.  

\begin{center}
\begin{tcolorbox}[skin=widget,
boxrule=1mm,
coltitle=black,
colframe=yellow!45!white,
colback=yellow!15!white,
width=(\linewidth),center,before=\hfill,after=\hfill]
\textbf{RQ8}: How does the API response structure impact the performance of LLMs on JSON processing tasks? \\
\textbf{Insight}: There is no clear correlation between the level of nesting or the endpoints and the performance of LLMs. 
\end{tcolorbox}
\end{center}

\section{Practitioner's Guidance}
The suitability of each processing approach depends on multiple factors. When tool responses are particularly large, appending them in a multi-turn setting will soon make the context run out (even for models with large context, 128K and above). Complex JSON structures such as their nesting depth or semantic similarity of their keys may also impacts this decision. 

% Some models are very chatty and don't follow instructions to restrict their verbosity; when the output of the tool will be forwarded to the chat interface, verbosity will not interfere with the system's ability to successfully accomplish a user's query. However, when the system needs to use the tool response to formulate the next tool call in the sequence, a verbose output can affect the execution of the next call. 
Answer generation results in verbose outputs compared to code generation, so the choice depends on whether the downstream usage can tolerate verbosity. Custom parsing of LLM output will almost always outperform a generic parser. However, it may not be feasible to implement a parser for each model, especially as models improve over time and new models are created. Thus, using a generic parser (based on widely used programming languages like Python) becomes desirable.

\subsection{Decision Tree}

\begin{figure}[htb]
    \centering
    \includegraphics[width=\linewidth]{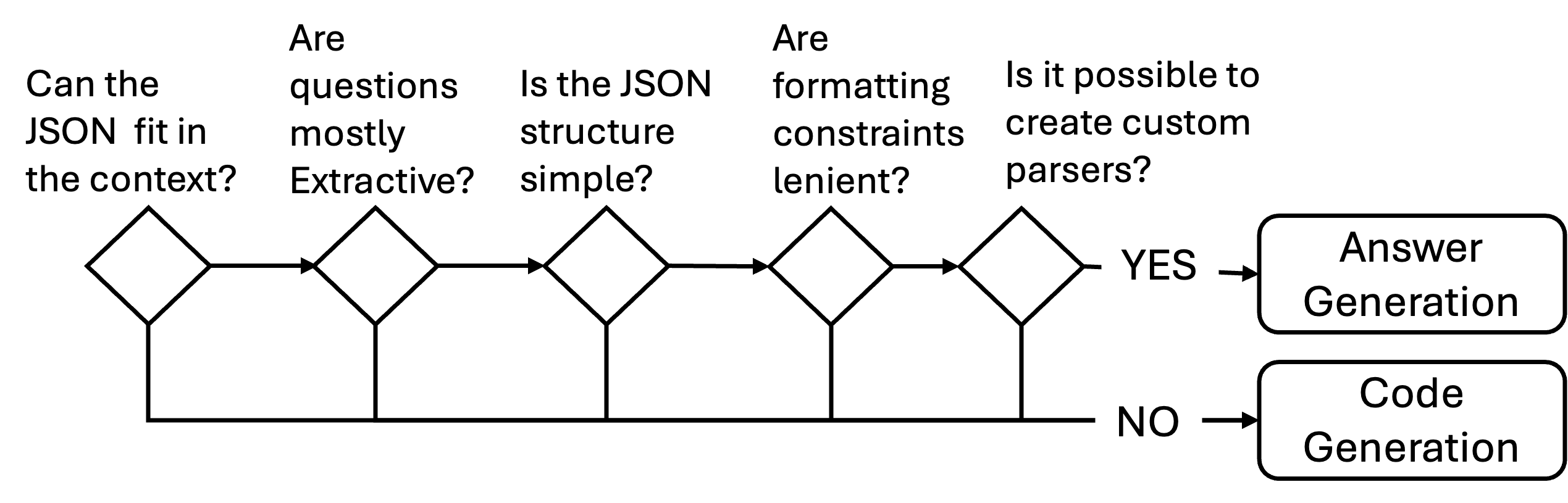}
    \caption{A practitioner's guidance diagram to inform the design decisions when using LLMs to process JSON responses. There is no one size fits all solution and model but there are some combinations that work better than others based on the tasks' characteristics. }
    \label{fig:guide}
\end{figure}

Based on the results we have presented in this paper, we derived a guidance diagram, shown in Figure \ref{fig:guide}, to help practitioners decide on the best approach to adopt. In general, code generation outperforms answer generation approaches under most circumstances. However, there are some circumstances under which answer generation is the better approach. Namely, when most questions are extractive on simpler JSONs where formatting issues in the LLM output can be tolerated. 

\subsection{Best Code Generation Model}
Furthermore, results show that GPT-OSS-120B was the best overall model for code generation. Thus, whenever possible, using that model to process JSON responses could lead to better results. 

\subsection{Sandboxing}
Finally, it is important to note that we cannot provide any guarantees that the code generated for parsing the responses is safe and trusted. As a precaution against any undesirable consequences from running this code, it would be in the best interest of a practitioner to ensure that execution only happens within a sandbox.

\section{Examples of Model Errors}
\label{appendix:error_analysis}

This section describes some of the error patterns we see from the model outputs. We took care of handling some of them in the \textit{exact match} metric calculation. For example, when the model is asked to return a comma separated list of some kind of values, the \textit{exact match} computation will handle duplicates and difference in the order of values compared to the ground truth.

\begin{figure*}[ht]
\begin{lstlisting}[basicstyle=\fontsize{7}{7}\selectfont\ttfamily, label={lst:error_nesting_structure}]
------------------------------------------------------------------------------------------------------------
Question : What is the arrival airport code for flight 1406?
------------------------------------------------------------------------------------------------------------
JSON Snippet (stripped down for clarity):
{
    "departureAirport": {
        "type": "AIRPORT",
        "code": "MUC",
        "name": "Munich Airport",
        "city": "MUC",
        "cityName": "Munich",
        "country": "DE",
        "countryName": "Germany"
    },
    "arrivalAirport": {
        "type": "AIRPORT",
        "code": "ARN",
        "name": "Stockholm Arlanda Airport",
        "city": "STO",
        "cityName": "Stockholm",
        "country": "SE",
        "countryName": "Sweden"
    },
    "departureTime": "2024-11-26T18:40:00",
    "arrivalTime": "2024-11-26T23:50:00",
    "legs": [
        {
            "departureTime": "2024-11-26T18:40:00",
            "arrivalTime": "2024-11-26T22:10:00",
            "departureAirport": {
                "type": "AIRPORT",
                "code": "MUC",
                "name": "Munich Airport",
                "city": "MUC",
                "cityName": "Munich",
                "country": "DE",
                "countryName": "Germany"
            },
            "arrivalAirport": {
                "type": "AIRPORT",
                "code": "HEL",
                "name": "Helsinki-Vantaa Airport",
                "city": "HEL",
                "cityName": "Helsinki",
                "country": "FI",
                "countryName": "Finland"
            },
            "cabinClass": "ECONOMY",
            "flightInfo": {
                "flightNumber": 1406,
                "planeType": "320",
                "carrierInfo": {
                    "operatingCarrier": "AY",
                    "marketingCarrier": "AY"
                }
            },
            "totalTime": 9000,
            "departureTerminal": "1"
        },
        {
            "departureTime": "2024-11-26T23:50:00",
            "arrivalTime": "2024-11-26T23:50:00",
            "departureAirport": {
                "type": "AIRPORT",
                "code": "HEL",
                "name": "Helsinki-Vantaa Airport",
                "city": "HEL",
                "cityName": "Helsinki",
                "country": "FI",
                "countryName": "Finland"
            },
            "arrivalAirport": {
                "type": "AIRPORT",
                "code": "ARN",
                "name": "Stockholm Arlanda Airport",
                "city": "STO",
                "cityName": "Stockholm",
                "country": "SE",
                "countryName": "Sweden"
            },
            "cabinClass": "ECONOMY",
            "flightInfo": {
                "flightNumber": 825,
                "planeType": "320",
                "carrierInfo": {
                    "operatingCarrier": "AY",
                    "marketingCarrier": "AY"
                }
            },
            "totalTime": 3600,
            "arrivalTerminal": "2"
        }
    ],
    "totalTime": 18600,
    "isAtolProtected": false,
    "showWarningDestinationAirport": false,
    "showWarningOriginAirport": false
}
------------------------------------------------------------------------------------------------------------
Ground Truth: HEL
Predicted: ARN
------------------------------------------------------------------------------------------------------------

\end{lstlisting}
\caption{Example of an error where the model confused the nesting structure}
\label{fig:error_nesting_structure}
% \vspace{-73.21692pt} % silences logs about float issue. other options?
\end{figure*}

\begin{enumerate}
    \item Comparing the average length of predicted responses to the length of the golden response (in terms of number of characters), some models were very verbose despite instructions in the prompt to only return the answer to the question. This was particularly pronounced for extractive questions and certain  models. For example, on about 500 samples where the average gold response had an average of 6 characters, answer generation DeepSeek-R1-distill-Llama-70B led to predictions of 1044 characters (on average). 
    For example, the question was: "What is the total number of available rooms of the kind Atura King Room - Free cancellation?"
    Model output: ""20
</think><think>
Alright, I need to figure out the total number of available rooms of the kind "Atura King Room - Free cancellation" based on the provided JSON data. 

First, I'll look through the "available" array in the JSON. Each object in this array represents a different room type. I'm specifically interested in the rooms where the "name" field is "Atura King Room - Free cancellation".

Looking at the first object in the "available" array, I see that the "name" is indeed "Atura King Room - Free cancellation". The "room\_count" for this room is 20. 

I'll check the other objects in the "available" array to see if there are any other entries with the same name. The second object has a different name, "Atura King Room - single occupancy - Free cancellation", so it doesn't match. The third object is another variation, and the fourth and fifth also have different names. 

Therefore, only the first object has the exact name we're looking for, and its "room\_count" is 20. There are no other objects with the same name, so the total number of available rooms of that specific type is 20.
</think>""

    \item Often, REST API responses have multiple keys that are semantically similar and it is confusing for LLMs to pick the information from the exact key. For example, the Get\_Room\_List\_With\_Availability endpoint from booking.com 
produces a response that has three different keys referring to the room name: name, room\_name, and name\_without\_policy. These have different values in them and the models return wrong values due to such confusion.
    \item As mentioned earlier, aggregation operation is harder with answer generation. For example, for the question "List available rooms with gross rate less than 112.89 USD. Output a comma separated list of room names." The models fail to do the comparison in the answer generation setting but such comparison is an easy and precise task for code.
    \item In addition to the answer to the question, models regurgitate the input JSON response or output reasoning or sometimes some ambiguous answers for some questions in answer generation. For example, for the question 'Provide the number of shoes which are available in Black colour.'
    The output from the model is:
    'Note: The answer is based on the provided JSON data. The product with ID "13857994671980346845" has a "Color" attribute with value "Black". 
    However, there is another product with ID "16782429352695976508" which has a "Color" attribute with value "Burgundy" and "Maroon" and also has a thumbnail with a product ID "16782429352695976508" which is the same as the product ID of the "Burgundy" and "Maroon" color variant. This product ID "16782429352695976508" is also present in the "Color" attribute of the product with ID "2345888757619488155" with value "Burgundy" and "Maroon". 
    But there is no product with ID "16782429352695976508" in the provided JSON data. 
    So, the answer is based on the product with ID "13857994671980346845" which has a "Color" attribute with value "Black".'

    \item As described earlier, models do not follow formatting instructions. For example, for a question where we just want the exact number as output and the prompt explicitly says do not output the unit, it outputs the unit (such as USD for prices, LB for weight).
    \item Code generation sometimes produced different logic which led to incorrect answers, partially attributed to ambiguous or convoluted natural language queries. 
\end{enumerate}

\end{document}